\def\year{2022}\relax
\documentclass[letterpaper]{article} %
\usepackage{aaai22}  %
\usepackage{times}  %
\usepackage{helvet}  %
\usepackage{courier}  %
\usepackage[hyphens]{url}  %
\usepackage{graphicx} %
\usepackage{natbib}  %
\usepackage{caption} %
\DeclareCaptionStyle{ruled}{labelfont=normalfont,labelsep=colon,strut=off} %
\usepackage{algorithm}
\usepackage{algorithmic}

\usepackage{newfloat}
\usepackage{listings}
\floatstyle{ruled}
\newfloat{listing}{tb}{lst}{}
\floatname{listing}{Listing}
\usepackage[colorlinks=true, linkcolor=red, urlcolor=blue, citecolor=blue, anchorcolor=blue]{hyperref}

\usepackage{booktabs}
\usepackage{multirow}
\usepackage[utf8]{inputenc} %
\usepackage{amsfonts}       %
\usepackage{nicefrac}       %
\usepackage{microtype}      %
\usepackage{lipsum}
\usepackage[dvipsnames]{xcolor}
\usepackage{subcaption}
\usepackage{tikz}
\usetikzlibrary{bayesnet}
\usepackage{centernot}
\DeclareMathAlphabet{\mathcal}{OMS}{cmsy}{m}{n}
\usepackage{setspace}
\usepackage{pdflscape}
\usepackage{colortbl}
\usepackage{textcomp}
\usepackage{float}
\usepackage{enumitem}
\usepackage{mathtools} %
\usepackage{amsthm}
\usepackage{thmtools, thm-restate}
\usepackage{cleveref}
\newcommand{\ra}[1]{\renewcommand{\arraystretch}{#1}}
\crefname{figure}{Fig.}{Figs.}
\crefname{definition}{Defn.}{Defns.}
\crefname{corollary}{Corollary}{Corollaries}
\crefname{proposition}{Prop.}{Props.}
\crefname{theorem}{Thm.}{Thms.}
\crefname{remark}{Remark}{Remarks}
\crefname{principle}{Principle}{Principles}
\crefname{lemma}{Lemma}{Lemmas}
\crefname{claim}{Claim}{Claims}
\crefname{table}{Tab.}{Tabs.}
\crefname{section}{\S}{\S\S}
\crefname{subsection}{\S}{\S\S}
\crefname{subsubsection}{\S}{\S\S}

\newcommand{\dec}{D}

\usepackage{notation}

\title{On the Fairness of Causal Algorithmic Recourse}
\author {
    Julius von K\"ugelgen,\textsuperscript{\rm 1,2}
    Amir-Hossein Karimi,\textsuperscript{\rm 1,3}
    Umang Bhatt,\textsuperscript{\rm 2}\\
    Isabel Valera,\textsuperscript{\rm 4}
    Adrian Weller,\textsuperscript{\rm 2,5}
    Bernhard Sch\"olkopf\textsuperscript{\rm 1}
}
\affiliations {
    \textsuperscript{\rm 1} Max Planck Institute for Intelligent Systems T\"ubingen\\
 \textsuperscript{\rm 2} University of Cambridge\\
  \textsuperscript{\rm 3} ETH Z\"urich\\
 \textsuperscript{\rm 4} Saarland University\\
 \textsuperscript{\rm 5} The Alan Turing Institute\\
    \texttt{jvk@tue.mpg.de, amir@tue.mpg.de, usb20@cam.ac.uk,\\ ivalera@cs.uni-saarland.de, aw665@cam.ac.uk, bs@tue.mpg.de\\}
}

\begin{document}

\maketitle

\begin{abstract}
Algorithmic fairness is typically studied from the perspective of \textit{predictions}. Instead, here we investigate fairness from the perspective of \textit{recourse} actions suggested to individuals to remedy
an unfavourable classification.
We propose two new fairness criteria at the group  and individual level, which---unlike prior work on equalising the average group-wise distance from the decision boundary---explicitly account for causal relationships between features, thereby capturing downstream effects of recourse actions performed in the physical world.
We explore how our criteria relate to others, such as counterfactual fairness, and show that fairness of recourse 
is complementary to fairness of prediction.
We study theoretically and empirically  how to enforce fair causal recourse by altering the classifier and perform a case study on the Adult dataset.  
Finally, we discuss whether
fairness violations in the data generating process revealed by our criteria may be better addressed by societal interventions 
as opposed to constraints on the classifier.
\end{abstract}

\section{Introduction}
\label{sec:introduction}

\emph{Algorithmic fairness} is concerned with uncovering and correcting for potentially discriminatory behavior of automated decision making systems~\citep{dwork2012fairness,zemel2013learning,hardt2016equality,chouldechova2017fair}.
Given a dataset comprising individuals from multiple legally protected groups (defined, e.g., based on age, sex, or ethnicity), and a binary classifier trained to predict a decision (e.g., whether they were approved for a credit card), most approaches to algorithmic fairness seek to quantify the level of unfairness according to a pre-defined (statistical or causal) criterion, and then aim to correct it by altering the classifier.
This notion of \textit{predictive fairness} typically considers the \textit{dataset as fixed}, and thus the \textit{individuals as unalterable}. 

\emph{Algorithmic recourse}, on the other hand, is concerned with offering recommendations to individuals, who were unfavourably treated by a decision-making system, to overcome their adverse situation~\citep{joshi2019towards,ustun2019actionable,sharma2019certifai,mahajan2019preserving,mothilal2020explaining,venkatasubramanian2020philosophical,karimi2020imperfect,karimi2020survey,karimi2020mint,upadhyay2021towards}.
For a given classifier and a negatively-classified individual, algorithmic recourse aims to identify which changes the individual could perform to flip the decision.
\looseness-1 Contrary to predictive fairness, recourse thus considers the \textit{classifier as fixed} but \textit{ascribes agency to the individual}.

\begin{figure}[t]
    \centering
    \includegraphics[width=\columnwidth, trim={0 0.0cm 0 0cm},clip]{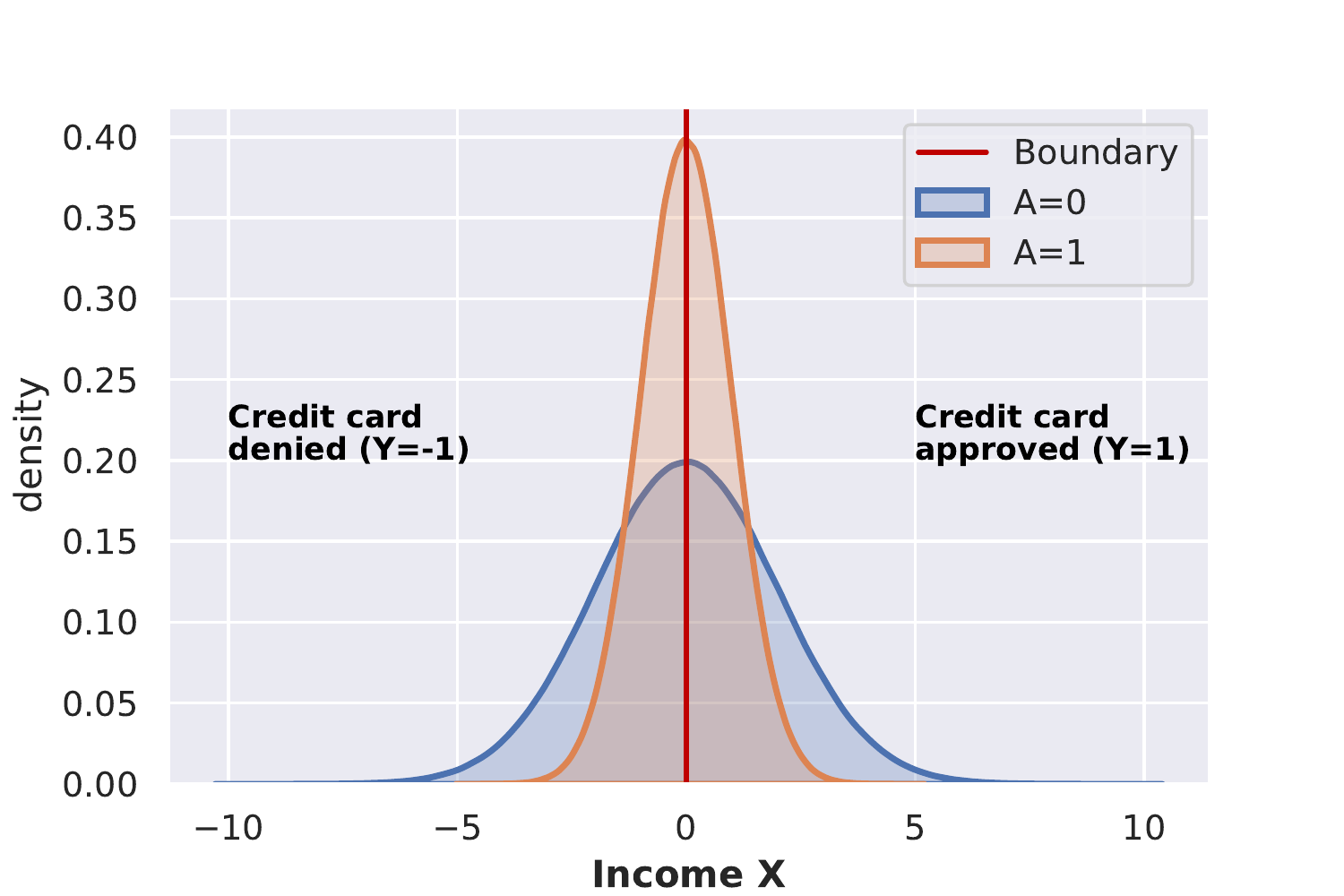}
    \caption{Example demonstrating the difference between fair \textit{prediction} and fair \textit{recourse}:
    here, only the variance of (centered income) $X$ differs across two protected groups $A \in \{0,1\}$, while the true and predicted label (whether an individual is approved for a credit card) are determined by $\sgn(X)$.
    This scenario would be considered fair from the perspective of \textit{prediction}, but the cost of \textit{recourse} (here, the distance to the decision boundary, set at $X=0$) is much larger for individuals in the blue group with $A=0$.
    }
    \label{fig:different_variances}
\end{figure}

Within machine learning (ML), fairness and recourse have mostly been considered in isolation and viewed as separate problems.
While recourse has been investigated in the presence of protected attributes---e.g., by comparing recourse actions 
(flipsets) 
suggested to otherwise similar male and female individuals~\cite{ustun2019actionable}, or comparing the aggregated cost of recourse 
(burden)
across different protected groups~\cite{sharma2019certifai}---its relation to fairness has only been studied informally, in the sense that differences in recourse have typically been understood as \textit{proxies of predictive unfairness}~\cite{karimi2020model}.
However, as we argue in the present work, recourse actually
constitutes an interesting fairness criterion \textit{in its own right} as it allows for the notions of agency and effort to be integrated into the study of fairness.

In fact, \textit{discriminatory recourse } \textit{does not imply predictive unfairness} (and is not implied by it either\footnote{Clearly, the \textit{average cost of recourse} across groups can be the \textit{same}, even if the \textit{proportion} of individuals which are classified as positive or negative is very \textit{different} across groups}).
To see this, consider the data shown in~\Cref{fig:different_variances}.
Suppose the feature $X$ represents the (centered) income of an individual from one of two sub-groups $A \in \{0,1\}$, distributed as $\mathcal{N}(0,1)$ and $\mathcal{N}(0,4)$, i.e., only the variances differ. 
Now consider a binary classifier $h(X) = \text{sign}(X)$ which perfectly predicts whether the individual is approved for a credit card (the true label $Y$)~\citep{barocas2020hidden}.
While this scenario satisfies several \textit{predictive fairness} criteria (e.g., demographic parity, equalised odds, calibration), the required increase in income for negatively-classified individuals to be approved for a credit card  (i.e., the effort required to achieve recourse) is much larger for the higher variance group. 
If individuals from one protected group need to work harder than ``similar'' ones from another group to achieve the same goal, this violates the concept of equal opportunity, a notion aiming for people to operate
on a level playing field~\cite{sep-equal-opportunity}.\footnote{This differs from the commonly-used purely predictive, statistical criterion of equal opportunity~\cite{hardt2016equality}.}
However, this type of unfairness is not captured by predictive notions which---in only distinguishing between (unalterable) worthy or unworthy individuals---do not consider the possibility for individuals to deliberately improve their situation by means of changes or interventions.

In this vein, \citet{gupta2019equalizing} recently introduced Equalizing Recourse, the first recourse-based and prediction-independent notion of fairness in ML.
They propose to measure recourse fairness in terms of the \textit{average group-wise distance to the decision boundary} for those getting a bad outcome, and show that this can be calibrated during classifier training.
However, this
formulation ignores that \textit{recourse is fundamentally a causal problem} since actions performed by individuals in the real-world to change their situation may have downstream effects%
~\citep{mahajan2019preserving,karimi2020mint,karimi2020imperfect,mothilal2020explaining}, cf.\,also~\citep{barocas2020hidden, wachter2017counterfactual,ustun2019actionable}.
By not reasoning about 
causal relations between features, the distance-based approach (i) does not accurately reflect the true (differences in) recourse cost, and (ii) is restricted to the classical prediction-centered approach of changing the classifier to address discriminatory recourse.

In the present work, we address both of these limitations.
First, by extending the idea of Equalizing Recourse to the minimal intervention-based framework of recourse~\cite{karimi2020mint}, we introduce \textit{causal} notions of fair recourse
which capture the true differences in recourse cost more faithfully 
if
features are not independently manipulable, as is generally the case.
Second, we argue that a causal model of the data generating process opens up a new route to fairness via \textit{societal interventions} in the form of changes to the underlying system. Such societal interventions may reflect common policies like subgroup-specific subsidies or tax breaks.
We highlight the following contributions:
\begin{itemize}[]
    \item we introduce a \textit{causal} version~(\cref{def:MINT-fair}) of Equalizing Recourse, as well as a stronger~(\cref{prop:group_level_insufficient_for_individual_level}) \textit{individual}-level criterion~(\cref{def:counterfactually-MINT-fair}) which we argue is more appropriate;
    \item we provide the first \textit{formal} study of the relation between fair prediction and fair recourse, and show that they are complementary notions which do not imply each other~(\cref{prop:CF_fairness_does_not_imply_fair_recourse});
    \item we establish sufficient conditions that allow for individually-fair causal recourse~(\cref{prop:nondescendants_imply_fair_recourse});
    \item we evaluate different fair recourse metrics for several classifiers~(\cref{sec:experiments_numerical}), verify our main results, and demonstrate that non-causal metrics misrepresent recourse unfairness;
    \item in a case study on the Adult dataset, we detect recourse discrimination at the group and individual level~(\cref{sec:experiments_adult}), demonstrating its relevance for real world settings;
    \item we propose societal interventions as an alternative to altering a classifier to address unfairness~(\cref{sec:social_interventions}).
\end{itemize}

\section{Preliminaries \& Background
}
\label{sec:background}
\paragraph{Notation.}
Let the random vector $\Xb=(X_1, ..., X_n)$ taking values $\xb=(x_1, ..., x_n)\in\Xcal=\Xcal_1\times ...  \times \Xcal_n\subseteq\RR^n$ denote
observed (non-protected) features. Let the random variable
$A$ taking values $a\in\Acal = \{1, \ldots, K\}$ for some $K\in\ZZ_{>1}$ denote a (legally) protected attribute/feature indicating which group each individual belongs to (based, e.g., on her age, sex, ethnicity, religion, etc).
And let $h:\Xcal\rightarrow \Ycal$ be a \textit{given} binary classifier with 
$Y\in\Ycal=\{\pm 1\}$ denoting the ground truth label (e.g., whether her credit card was approved).
We observe a dataset $\Dcal=\{\vb^i\}_{i=1}^N$ of i.i.d.\ observations of the random variable $\Vb=(\Xb,A)$ with~$\vb^i:=(\xb^i,a^i)$.\footnote{We use $\vb$ when there is an explicit distinction between the protected attribute and other features (in the context of fairness) and $\xb$ otherwise (in the context of explainability).}
\paragraph{Counterfactual Explanations.}
A common framework for  explaining decisions made by (black-box) ML models is that of counterfactual explanations~\cite[\CE ;][]{wachter2017counterfactual}.
A \CE\ is a closest
feature vector on the other side of the decision boundary. 
Given a distance $d:\Xcal\times\Xcal\rightarrow\RR^+$, a \CE\ for an individual $\xb^\Ftt$ who obtained an unfavourable prediction, $h(\xb^\Ftt)=-1$, is  defined as a solution to:%
\begin{equation}
\label{eq:counterfactual_explanation}
\min_{\xb\in\Xcal}\, d(\xb, \xb^\Ftt)
\quad \quad 
\text{subject to} 
\quad \quad 
 h(\xb)=1.
\end{equation}%
While \CEs\ are useful to \textit{understand the behaviour of a classifier}, they do not generally lead to \textit{actionable recommendations}: they inform an individual of where she should be to obtain a more favourable prediction, but they may not suggest \textit{feasible} changes she could perform to get there.

\newcommand{\colorIMF}{RedOrange}
\newcommand{\colorCAU}{ForestGreen}
\newcommand{\xshift}{3em}
\newcommand{\yshift}{2.5em}
\begin{figure}[t]
    \begin{subfigure}[b]{0.5\columnwidth}
        \centering
        \begin{tikzpicture}
            \centering
            \node (A) [latent] {$A$};
            \node (X_1) [latent, below=of A, xshift=-\xshift, yshift=\yshift, fill=\colorIMF, fill opacity=0.5, text opacity=1] {$X_1$};
            \node (X_2) [latent, below=of A, xshift=\xshift, yshift=\yshift, fill=\colorIMF, fill opacity=0.5, text opacity=1] {$X_2$};
            \node (X_3) [latent, below=of X_1, xshift=\xshift, yshift=\yshift, fill=\colorIMF, fill opacity=0.5, text opacity=1] {$X_3$};
            \edge[]{A}{X_1, X_2, X_3};
        \end{tikzpicture} 
        \caption{IMF assumption}
        \label{fig:causal_graph_IMF}
    \end{subfigure}%
    \begin{subfigure}[b]{0.5\columnwidth}
        \centering
        \begin{tikzpicture}
            \centering
            \node (A) [latent] {$A$};
            \node (X_1) [latent, below=of A, xshift=-\xshift, yshift=\yshift, fill=\colorCAU, fill opacity=0.5, text opacity=1] {$X_1$};
            \node (X_2) [latent, below=of A, xshift=\xshift, yshift=\yshift, fill=\colorCAU, fill opacity=0.5, text opacity=1] {$X_2$};
            \node (X_3) [latent, below=of X_1, xshift=\xshift, yshift=\yshift, fill=\colorCAU, fill opacity=0.5, text opacity=1] {$X_3$};
            \edge[]{A}{X_1, X_2, X_3};
            \edge[color=\colorCAU]{X_1}{X_2};
            \edge[color=\colorCAU]{X_1, X_2}{X_3};
        \end{tikzpicture}    
        \caption{Causal view}
        \label{fig:causal_graph_MINT}
    \end{subfigure}%
    \caption{%
    \text{(a)} The framework underlying counterfactual explanations and distance-based recourse treats $X_i$ as \textcolor{\colorIMF}{independently manipulable features (IMF)}.
    In a fairness context, this means that the $X_i$ may depend on the protected attribute $A$ (and potentially other unobserved factors) but do not causally influence each other.
    \text{(b)}~The present work considers a generalisation the IMF assumption by allowing for \textcolor{\colorCAU}{causal influences between the $X_i$}, thus modeling the 
    \textcolor{\colorCAU}{downstream effects} of changing some features on others. This causal approach allows us to more accurately quantify recourse unfairness in real-world settings where the IMF assumption is typically violated. It also provides a framework for studying alternative routes to achieve fair recourse beyond changing the classifier.
    }
  \label{fig:causal_graphs}
\end{figure}
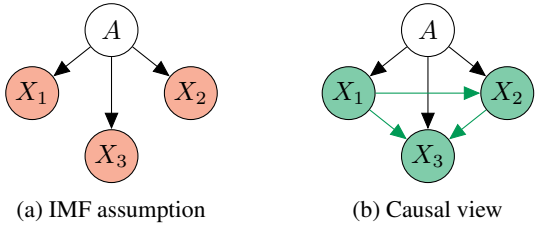

\paragraph{Recourse with Independently-Manipulable Features.}
\citet{ustun2019actionable} refer to a person's  ability to change the decision of a model by altering actionable variables as \textit{recourse} 
and propose to solve
\begin{equation}
\begin{aligned}
\label{eq:independent_world_recourse}
    \min_{\bm\delta\in\Fcal(\xF)}\, c(\bm\delta; \xF)
    \quad 
    \text{subject to}
    \quad 
    h(\xF+\bm\delta)=1
\end{aligned}
\end{equation}
where $\Fcal(\xF)$ is a set of feasible change vectors and $c(\cdot; \xF)$ is a cost function defined over these actions, both of which may depend on the individual.\footnote{For simplicity, \eqref{eq:independent_world_recourse}~assumes that all $X_i$ are continuous; we do not make this assumption in the remainder of the present work. }
As pointed out by~\citet{karimi2020mint},
\eqref{eq:independent_world_recourse} 
implicitly treats features as manipulable independently of each other (see~\cref{fig:causal_graph_IMF}) and does not account for causal relations that may exist between them (see~\cref{fig:causal_graph_MINT}): while allowing feasibility constraints on actions, 
variables which are not acted-upon ($\delta_i=0$) are assumed to remain unchanged.
We refer to this 
as the
\textit{independently-manipulable features}
(IMF) assumption.
While the IMF-view may be appropriate when only analysing the behaviour of a classifier,
it falls short of capturing effects of interventions performed in the real world, as is the case in actionable recourse; e.g., an increase in income will likely also positively affect the individual's savings balance.
As a consequence, 
\eqref{eq:independent_world_recourse} only guarantees recourse if the acted-upon variables have no causal effect on the remaining variables~\citep{karimi2020mint}.

\paragraph{Structural Causal Models.}
A structural causal model (\SCM)~\citep{pearl2009causality, peters2017elements} over
observed variables $\Vb=\{V_i\}_{i=1}^n$ is a pair~$\Mcal=(\Sb, \PP_\Ub)$, where the structural equations $\Sb$  are a set of assignments $\Sb=\{V_i:=f_i(\PA_i, U_i)\}_{i=1}^{n}$, which compute each $V_i$ as a deterministic function $f_i$ of its direct causes (causal parents) $\PA_i\subseteq\Vb\setminus V_i$ and an unobserved variable~$U_i$. 
In this work, we make the common assumption that the distribution $\PP_\Ub$ factorises over the latent $\Ub=\{U_i\}_{i=1}^n$, meaning that there is no unobserved confounding (causal sufficiency).
If the causal graph $\Gcal$ associated with $\Mcal$ (obtained by drawing a directed edge from each variable in $\PA_i$ to $V_i$, see~\cref{fig:causal_graphs} for examples) is acyclic, $\Mcal$ induces a unique ``observational'' distribution over $\Vb$, defined as the push forward of $\PP_{\Ub}$ via~$\Sb$.

SCMs can be used to model the effect of \textit{interventions}: external manipulations to the system that change the generative process (i.e., the structural assignments) of a subset of variables~$\Vb_\Ical\subseteq\Vb$, e.g., by fixing their value to a constant~$\thetaI$.
Such (atomic) interventions are denoted using Pearl's $do$-operator by $do(\Vb_\Ical:=\thetaI)$, or $do(\thetaI)$ for short.
Interventional distributions are obtained from $\Mcal$ by replacing the structural equations $\{V_i:=f_i(\PA_i, U_i)\}_{i\in\Ical}$ by their new assignments $\{V_i:=\theta_i\}_{i\in\Ical}$ to obtain the modified structural equations $\Sb^{do(\thetaI)}$ and then computing the distribution induced by the interventional SCM $\Mcal^{do(\thetaI)}=(\Sb^{do(\thetaI)},\PP_\Ub)$, i.e., the push-forward of $\PP_\Ub$ via $\Sb^{do(\thetaI)}$.

Similarly, \SCMs allow reasoning about (structural) \textit{counterfactuals}: statements about interventions performed in a hypothetical world where all unobserved noise terms $\Ub$ are kept unchanged and fixed to their factual value $\ub^\Ftt$.
The counterfactual distribution for a hypothetical intervention $do(\thetaI)$ given a factual observation $\vb^\Ftt$, denoted $\vb_{\thetaI}(\uF)$, can be obtained from $\Mcal$ using a three step procedure: first, inferring the posterior distribution over the unobserved variables $\PP_{\Ub|\vb^\Ftt}$ (\textit{abduction}); second, replacing some of the structural equations as in the interventional case (\textit{action}); third, computing the distribution induced by the counterfactual SCM $\Mcal^{do(\thetaI)|\vb^\Ftt}=(\Sb^{do(\thetaI)},\PP_{\Ub|\vb^\Ftt})$ (\textit{prediction}).
\paragraph{Causal Recourse.}
To capture causal relations between features, \citet{karimi2020mint} propose to approach the actionable recourse task within the framework of \SCMs and to shift the focus from nearest \CEs\ to minimal interventions, leading to the optimisation problem
\begin{equation}
\begin{aligned}
\label{eq:causal_recourse}
    \min_{\thetaI\in\Fcal(\xF)}\, c(\thetaI;\xF)
    \quad
    \text{subj.\ to}
    \quad
    h(\xb_{\thetaI}(\uF))=1,
\end{aligned}
\end{equation}
where $\xb_{\thetaI}(\uF)$ denotes the ``counterfactual twin'' of $\xF$ had $\Xb_\Ical$ been $\thetaI$.\footnote{For an interventional notion of recourse related to conditional average treatment effects (CATE) for a specific subpopulation, see~\cite{karimi2020imperfect}; in the present work, we focus on the individualised counterfactual notion of causal recourse.}
In practice, the \SCM\ is unknown and needs to be inferred from data based on additional (domain-specific) assumptions, leading to probabilistic versions of~\eqref{eq:causal_recourse} which aim to find
actions that achieve recourse with high probability~\citep{karimi2020imperfect}.
If the IMF assumptions holds (i.e.,  the set of descendants of all actionable variables is empty), then \eqref{eq:causal_recourse}~reduces to IMF recourse%
~\eqref{eq:independent_world_recourse} as a special case.

\paragraph{Algorithmic and Counterfactual Fairness.}
While there are many statistical notions of fairness \citep{zafar2017fairness, zafar2017constraints}, these are sometimes mutually incompatible~\citep{chouldechova2017fair}, and it has been argued that discrimination, at its heart, corresponds to a (direct or indirect) causal influence of a protected attribute on the prediction, thus making fairness a fundamentally causal problem~\citep{kilbertus2017avoiding,russell2017worlds, loftus2018causal, zhang2018equality, zhang2018fairness, nabi2018fair, nabi2019learning, chiappa2019path, salimi2019interventional,wu2019pc}.
Of particular interest to our work is the notion of \textit{counterfactual fairness} introduced by~\citet{kusner2017counterfactual} 
which calls a (probabilistic) classifier $h$ over $\Vb=\Xb\cup A$ counterfactually fair if it satisfies 
\begin{equation*}
    h(\vF)=h(\vb_{a}(\uF)), \forall a\in\Acal,
    \vF=(\xF,\aF)\in\Xcal\times\Acal,
\end{equation*}
where $\vb_{a}(\uF)$ denotes the ``counterfactual twin'' of $\vF$ had the attribute been $a$ instead of $\aF$. 
\paragraph{Equalizing Recourse Across Groups.}
The main focus of this paper is the \textit{fairness of recourse actions} which, to the best of our knowledge,  was studied for the first time by \citet{gupta2019equalizing}. %
They advocate for equalizing the average cost of recourse across protected groups and to incorporate this as a constraint when training a classifier. 
Taking a distance-based approach in line with \CEs, they define the cost of recourse for $\xF$ with $h(\xF)=-1$ as the minimum achieved in \eqref{eq:counterfactual_explanation}:
\begin{equation}
\label{eq:distance-based-recourse-cost}
r^\IW(\xF)=\min_{\xb\in\Xcal}\, d(\xF, \xb) \quad \text{subj.\ to} \quad h(\xb)=1,
\end{equation}
which is equivalent to IMF-recourse \eqref{eq:independent_world_recourse}
if $c(\bm\delta; \xF)=d(\xF+\bm\delta,\xF)$ is chosen as cost function.
Defining the  protected subgroups,
    $G_a=\{\vb^i\in\Dcal:a^i=a\}$,
    and    
    $G_a^-=\{\vb\in G_a: h(\vb)=-1\}$,
the group-level cost of recourse (here, the average distance to the decision boundary) is then given by,
\begin{equation}
\label{eq:group-level-cost}
    \textstyle
    r^\IW(G_a^-)=\frac{1}{|G_a^-|}
    \sum_{\vb^i\in G_a^-}r^\IW(\xb^i).
\end{equation}
The idea of \textit{Equalizing Recourse} across groups~\cite{gupta2019equalizing} can then be summarised as follows.%
\begin{definition}[Group-level fair IMF-recourse, from~\cite{gupta2019equalizing}]
\label{def:IW-fair}
The group-level unfairness of \emph{recourse with independently-manipulable features} (IMF)
for a dataset $\Dcal$, classifier $h$, and distance metric $d$
is:%
\begin{equation*}
    \Delta_\mathbf{\mathrm{dist}}(\Dcal, h, d):=\max_{a,a'\in\Acal}\left|
    r^\IW(G_a^-) - r^\IW(G_{a'}^-)
    \right|
    .
\end{equation*}%
Recourse for $(\Dcal, h, d)$ is 
``group IMF-fair'' if~$\Delta_\mathbf{\mathrm{dist}}
=0$.
\end{definition}%

\section{Fair \textit{Causal} Recourse}
\label{sec:fair_causal_recourse}

Since~\cref{def:IW-fair}
rests on the IMF assumption,
it ignores causal relationships between variables, fails to account for downstream effects of  actions on other relevant features, and thus generally incorrectly estimates the true cost of recourse.
We argue that recourse-based fairness considerations 
should rest on a causal model that captures the effect of interventions performed in the physical world where features are often 
causally related to each other.
We therefore consider an \SCM\ $\Mcal$ over $\Vb=(\Xb,A)$ to model causal relationships between the protected attribute and the remaining features.

\subsection{Group-Level Fair Causal Recourse}
\label{sec:group-level-causally-fair-recourse}
\Cref{def:IW-fair} can be adapted to the causal (CAU)
recourse framework~\eqref{eq:causal_recourse} by replacing the minimum distance in~\eqref{eq:distance-based-recourse-cost} with the cost of recourse
within a causal model, i.e., the minimum achieved in~\eqref{eq:causal_recourse}:%
\begin{equation*}
\begin{aligned}
\label{eq:causal-recourse-cost}
r^\MINT(\vF)=\min_{\thetaI\in\Theta(\vF)}\, c(\thetaI;\vF)
\quad
\text{subj.\ to}
\quad h(\vb_{\thetaI}(\uF))=1,
\end{aligned}
\end{equation*}
where we recall that the constraint  $h(\vb_{\thetaI}(\uF))=1$ ensures that the counterfactual twin of $\vF$ in $\Mcal$ falls on the favourable side of the classifier.
Let $r^\MINT(G_a^-)$ be the average of $r^\MINT(\vF)$ across $G_a^-$, analogously to~\eqref{eq:group-level-cost}.
We can then define group-level fair causal recourse as follows.%
\begin{definition}[Group-level fair causal recourse] 
\label{def:MINT-fair}
The group-level unfairness of \emph{causal} (CAU) recourse 
for a dataset $\Dcal$, classifier $h$, and cost function $c$ \textit{w.r.t.\ an SCM} $\Mcal$
is given by:
\begin{equation*}
    \Delta_\mathbf{\mathrm{cost}}(\Dcal, h, c, \Mcal):=\max_{a,a'\in\Acal}\left|
    r^\MINT(G_a^-) - r^\MINT(G_{a'}^-)
    \right|
    .
\end{equation*}
Recourse for $(\Dcal, h, c, \Mcal)$ 
is ``group CAU-fair'' if~$\Delta_\mathbf{\mathrm{cost}}
=0$.
\end{definition}%
While~\cref{def:IW-fair} is agnostic to the  (causal) generative process of the data (note the absence of a reference SCM $\Mcal$ from~\cref{def:IW-fair}),
~\cref{def:MINT-fair}
takes causal relationships between features into account when calculating the cost of recourse. 
It thus captures the effect of actions and the necessary cost of recourse more faithfully when the IMF-assumption is violated, as is realistic for most applications.

A shortcoming of both~\cref{def:IW-fair,def:MINT-fair} is that they are group-level definitions, i.e., they only consider the \textit{average} cost of recourse across all individuals sharing the same protected attribute.
However, it has been argued from causal~\citep{chiappa2019path,wu2019pc%
} and non-causal~\citep{dwork2012fairness} perspectives that fairness is fundamentally an individual-level concept:\footnote{After all, it is not much consolation for an individual who was unfairly given an unfavourable prediction to find out that other members of the same group were treated more favourably}
group-level fairness still allows for unfairness at the level of the individual, provided that positive and negative discrimination cancel out across the group.
\looseness-1 This is one motivation behind counterfactual fairness~\citep{kusner2017counterfactual}: a decision is considered fair at the individual level if it would not have changed, had the individual belonged to a different protected group.

\subsection{Individually Fair Causal Recourse}
\label{sec:individualised-causally-fair-recourse}
Inspired by counterfactual fairness~\citep{kusner2017counterfactual},
we propose that (causal) recourse may be considered fair at the level of the individual if the cost of recourse would have been the same had the individual belonged to a different protected  group, i.e., under a counterfactual change to~$A$.%
\begin{definition}[Individually
fair causal recourse]
\label{def:counterfactually-MINT-fair}
The individual-level unfairness of \emph{causal} recourse 
for a dataset $\Dcal$, classifier $h$, and cost function $c$ \textit{w.r.t.\ an SCM} $\Mcal$
is 
\begin{equation*}
    \Delta_\mathrm{ind}(\Dcal, h, c, \Mcal)
    :=
    \max_{a\in\Acal; \vF\in\Dcal}
    \left|
    r^\MINT(\vF) - r^\MINT(\vb_a(\uF))
    \right|
\end{equation*}
Recourse
is ``individually CAU-fair'' if~$\Delta_\mathrm{ind}
=0$.
\end{definition}%
This is a stronger notion in the sense that it is possible to satisfy both group IMF-fair~(\cref{def:IW-fair}) and group CAU-fair recourse~(\cref{def:MINT-fair}), without satisfying~\cref{def:counterfactually-MINT-fair}:
\begin{proposition}
\label{prop:group_level_insufficient_for_individual_level}
Neither of the group-level notions of fair recourse (\cref{def:IW-fair} and \cref{def:MINT-fair}) are sufficient conditions for individually CAU-fair recourse (\cref{def:counterfactually-MINT-fair}), i.e.,
\begin{align*}
    \text{Group IMF-fair} &\centernot\implies \text{Individually CAU-fair}. 
    \\
    \text{Group CAU-fair} &\centernot\implies \text{Individually CAU-fair}.
\end{align*}%
\begin{proof}
 A counterexample is given by the following combination of SCM and classifier
\begin{align*}
    A&:=U_A, 
   \\
    X&:=AU_X+(1-A)(1-U_X), 
    \\ 
    U_A, U_X&\sim \mathrm{Bernoulli}(0.5),
    \\ 
    Y&:=h(X)=\sgn(X-0.5).
\end{align*}
We have $\PP_{X|A=0}=\PP_{X|A=1}=\mathrm{Bern}(0.5)$, so the distance to the boundary at $X=0.5$ is the same across groups.
The criterion for ``group IMF-fair'' recourse~(\cref{def:IW-fair}) is thus satisfied.

Since protected attributes are generally immutable (thus making any recourse actions involving changes to $A$ infeasible) and since there is only a single feature in this example (so that causal downstream effects on descendant features can be ignored), the distance between the factual and counterfactual value of $X$ is a reasonable choice of cost function also for causal recourse.
In this case, $(\Dcal, h, \Mcal)$ also satisfies group-level CAU-fair recourse (\cref{def:MINT-fair}).

However, for all $\vF=(x^\texttt{F}, a^\texttt{F})$ and any $a\neq a^\texttt{F}$,  we have $h(x^\texttt{F})\neq h(x_{a}(u_X^\texttt{F}))=1-h(x^\texttt{F})$, so it is maximally unfair at the individual level:
for any individual,
the cost of recourse would have been zero had the protected attribute been different, as the prediction would have flipped.
\end{proof}
\end{proposition}%

\subsection{Relation to Counterfactual Fairness}
\label{sec:relation_to_CF_fairness}
The classifier $h$ used in the proof of~\cref{prop:group_level_insufficient_for_individual_level}
is \emph{not} counterfactually fair.
This suggests to investigate their relation more closely:
\textit{does a counterfactually fair classifier imply fair (causal) recourse?} 
The answer is no.
\begin{proposition}
\label{prop:CF_fairness_does_not_imply_fair_recourse}
Counterfactual fairness is insufficient for any of the three notions of fair recourse:
\begin{align*}
    h \,\,\text{counterfactually fair}  &\centernot\implies \text{Group IMF-fair}\\
    h \,\,\text{counterfactually fair}  &\centernot\implies \text{Group CAU-fair}\\
    h \,\,\text{counterfactually fair}  &\centernot\implies \text{Individually CAU-fair}
\end{align*}%
\begin{proof}
A counterexample is given by the following combination of SCM and classifier:
\begin{equation}
\begin{aligned}
\label{eq:generative_process_variance_example}
    A&:=U_A,  &U_A&\sim \text{Bernoulli}(0.5),
    \\
    X&:=(2-A)U_X,     &U_X&\sim \Ncal(0,1),\\
    Y&:=h(X)=\sgn(X)
\end{aligned}
\end{equation}
which we used to generate~\Cref{fig:different_variances}.
As $\sgn(X)=\sgn(U_X)$, and $U_X$ is assumed fixed when reasoning about a counterfactual change of~$A$, $h$ is counterfactually fair.

However, $\PP_{X|A=0}=\Ncal(0,4)$ and $\PP_{X|A=1}=\Ncal(0,1)$, so the distance to the boundary 
(which is a reasonable cost for \MINT-recourse in this one-variable toy example)
differs at the group level.
Moreover, $X$ either doubles or halves when counterfactually changing $A$.
\end{proof}
\end{proposition}
\begin{remark}
An important characteristic of the counterexample used in the proof of~\cref{prop:CF_fairness_does_not_imply_fair_recourse} is that 
$h$ is \textit{deterministic}, which makes it possible that $h$ is counterfactually fair, even though it depends on a descendant of $A$. 
This is generally not be the case if $h$ is \textit{probabilistic} (e.g., a logistic regression), $h:\Xcal\rightarrow [0,1]$, \looseness-1 so that the probability of a positive classification decreases with the distance from the decision boundary.
\end{remark}

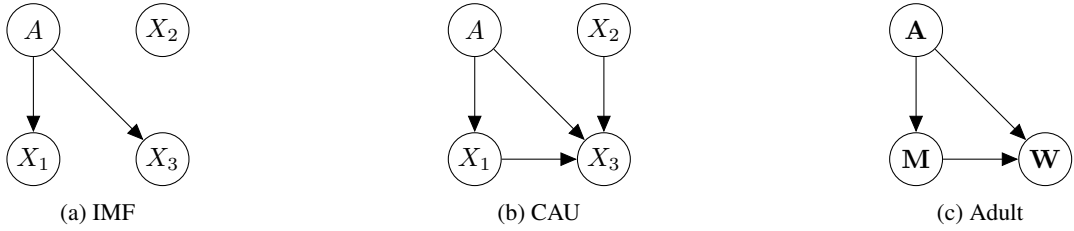
\begin{figure*}[t]
    \begin{subfigure}[b]{0.33\textwidth}
        \centering
        \begin{tikzpicture}
            \centering
            \node (A) [latent] {$A$};
            \node (X_1) [latent, below=of A] {$X_1$};
            \node (X_2) [latent, right=of A] {$X_2$};
            \node (X_3) [latent, below=of X_2] {$X_3$};
            \edge[]{A}{X_1, X_3};
        \end{tikzpicture} 
        \caption{IMF}
        \label{fig:IMF}
    \end{subfigure}%
    \begin{subfigure}[b]{0.33\textwidth}
        \centering
        \begin{tikzpicture}
             \centering
            \node (A) [latent] {$A$};
            \node (X_1) [latent, below=of A] {$X_1$};
            \node (X_2) [latent, right=of A] {$X_2$};
            \node (X_3) [latent, below=of X_2] {$X_3$};
            \edge[]{A}{X_1};
            \edge[]{A, X_1, X_2}{X_3};
        \end{tikzpicture}    
        \caption{CAU}
        \label{fig:CAU}
    \end{subfigure}%
    \begin{subfigure}[b]{0.33\textwidth}
        \centering
        \begin{tikzpicture}
             \centering
            \node (A) [latent] {$\mathbf{A}$};
            \node (M) [latent, below=of A] {$\mathbf{M}$};
            \node (W) [latent, right=of M] {$\mathbf{W}$};
            \edge[]{A}{M};
            \edge[]{A, M}{W};
        \end{tikzpicture}    
        \caption{Adult}
        \label{fig:adult}
    \end{subfigure}%
    \caption{(a) \& (b) Causal graphs used in~\cref{sec:experiments_numerical}. (c) The (assumed) causal graph (from \citet{chiappa2019path, nabi2018fair}) used for the Adult dataset~\cite{lichman2013uci}; $\mathbf{A}$ denotes the three protected attributes \{sex, age, nationality\}; $\mathbf{M}$ denotes \{marital status, education level\}; and $\mathbf{W}$ corresponds to \{working class, occupation, hrs per week\}. Here, we show the coarse-grained causal graph for simplicity. In practice, we model each node separately. For example, the single arrow from $A$ to $M$ actually corresponds to six directed edges, one from each feature in $\mathbf{A}$ to each feature in~$\mathbf{M}$.}
    \label{fig:synthetic_data_graph}
\end{figure*}

\subsection{Achieving Fair Causal Recourse 
}
\label{sec:achieving_recourse}

\paragraph{Constrained Optimisation.}
A first approach is to explicitly take constraints on the (group or individual level) fairness of causal recourse into account when training a classifier, as implemented for non-causal recourse under the IMF assumption by~\citet{gupta2019equalizing}.
Herein we can control the potential trade-off between accuracy and fairness with a hyperparameter.
However, the optimisation problem in~\eqref{eq:causal_recourse}
involves optimising over the combinatorial space of intervention targets $\Ical\subseteq\{1,...,n\}$, so 
it is unclear whether fairness of causal recourse may easily be included as a differentiable constraint.

\paragraph{Restricting the Classifier Inputs.}
\label{sec:restricted_inputs}
An approach 
that only requires \textit{qualitative} knowledge in form of the causal graph (but not a fully-specified \SCM), is to
restrict the set of input features to 
the classifier 
to only 
contain non-descendants of the protected attribute. In this case, and subject to some additional assumptions stated in more detail below, individually fair causal recourse can be guaranteed.
\begin{proposition}
\label{prop:nondescendants_imply_fair_recourse}
Assume $h$ only depends on a subset $\Tilde{\Xb}\subseteq\Vb\setminus(A\cup \mathrm{desc}(A))$ which are non-descendants of $A$ in $\Mcal$; and that the set of feasible actions 
and their cost  
remain the same under a counterfactual change of $A$, $\Fcal(\vF)=\Fcal(\vb_a(\uF))$ and $c(\cdot\,; \vF)=c(\cdot\,; \vb_a(\uF))$ $\forall a\in\Acal, \vF\in\Dcal$.
Then recourse for $(\Dcal, h, c, \Mcal)$ is ``individually CAU-fair''.
\begin{proof}
According to~\Cref{def:counterfactually-MINT-fair}, it suffices to show that 
\begin{equation}
r^\MINT(\vF) = r^\MINT(\vb_a(\uF)), \quad \quad \forall a\in\Acal, \vF\in\Dcal.
\end{equation}

Substituting our assumptions in the definition of $r^\MINT$ from~\cref{eq:causal-recourse-cost}, we obtain:
\begin{align*}
    r^\MINT(\vF)&=\min_{\thetaI\in\Fcal(\vF)}\, c(\thetaI;\vF)
    \quad 
    \text{s.t.} 
    \quad  h(\tilde{\xb}_{\thetaI}(\uF))=1,
\\
    r^\MINT(\vb_a(\uF))&=\min_{\thetaI\in\Fcal(\vF)}\, c(\thetaI;\vF)
    \quad
    \text{s.t.} 
    \quad h(\tilde{\xb}_{\thetaI, a}(\uF))=1.
\end{align*}
It remains to show that 
\begin{equation*}
\tilde{\xb}_{\thetaI, a}(\uF)=\tilde{\xb}_{\thetaI}(\uF),
\quad \quad \forall \thetaI\in\Fcal(\vF), a\in\Acal
\end{equation*}
which follows from applying do-calculus~\citep{pearl2009causality}
since $\Tilde{\Xb}$ does not contain any descendants of $A$ by assumption, and is thus not influenced by counterfactual changes to $A$.
\end{proof}
\end{proposition}

The assumption of~\Cref{prop:nondescendants_imply_fair_recourse} that both the set of feasible actions $\Fcal(\vF)$ and the cost function $c(\cdot\,;\vF)$ remain the same under a counterfactual change to the protected attribute may not always hold.
For example, if a protected group were precluded (by law) or discouraged from performing certain recourse actions such as taking on a particular job or applying for a certification, that would constitute such a violation due to a separate source of discrimination.

Moreover, since protected attributes usually represent socio-demographic features (e.g., age, gender, ethnicity, etc), they often appear as root nodes in the causal graph and have downstream effects on numerous other features. 
Forcing the classifier to only consider non-descendants of $A$ as inputs, as in~\Cref{prop:nondescendants_imply_fair_recourse}, can therefore lead to a 
drop in accuracy which can be a 
restriction~\cite{wu2019counterfactual}.

\paragraph{Abduction / Representation Learning.}
\label{sec:abduction}
We have shown that considering only non-descendants of $A$ is a way to achieve individually CAU-fair recourse.
In particular, this also applies to the unobserved variables $\Ub$ which are, by definition, not descendants of any observed variables.
This suggests to use $U_i$ in place of any descendants $X_i$ of $A$ when training the classifier---in a way, $U_i$ can be seen as a ``fair representation'' of $X_i$ since it is an exogenous 
component that is not due to $A$. 
However, as $\Ub$ is unobserved, it needs to be inferred from the observed $\vF$, corresponding to the abduction step of counterfactual reasoning.
Great care needs to be taken in learning such a representation in terms of the (fair) background variables as (untestable) counterfactual assumptions are required~\cite[\textsection~4.1]{kusner2017counterfactual}.

\begin{table*}[t]
    \ra{1.5}
    \centering
    \resizebox{\textwidth}{!}{
    \begin{tabular}{l  c c c c | c c c c | c c c c || c c c c | c c c c | c c c c}
         \toprule
         \multirow{3}{*}{\textbf{Classifier}}
         & \multicolumn{12}{c}{\textbf{GT labels from \textit{linear} log. reg. $\rightarrow$ using \textit{linear} kernel / \textit{linear} log. reg. }} 
         & \multicolumn{12}{c}{\textbf{GT labels from \textit{nonlinear} log. reg. $\rightarrow$ using \textit{polynomial} kernel / \textit{nonlinear} log. reg. }}
         \\
         \cmidrule(r){2-13} \cmidrule(r){14-25}
         & \multicolumn{4}{c}{\textbf{IMF}} 
         & \multicolumn{4}{c}{\textbf{CAU-LIN}}
         & \multicolumn{4}{c}{\textbf{CAU-ANM}}
         & \multicolumn{4}{c}{\textbf{IMF}} 
         & \multicolumn{4}{c}{\textbf{CAU-LIN}}
         & \multicolumn{4}{c}{\textbf{CAU-ANM}}
         \\
         \cmidrule(r){2-5} \cmidrule(r){6-9} \cmidrule(r) {10-13}
         \cmidrule(r){14-17} \cmidrule(r){18-21} \cmidrule(r) {22-25}
         & \textbf{Acc}
         & $\Delta_\textbf{dist}$
         & $\Delta_\textbf{cost}$
         & $\Delta_\textbf{ind}$
         & \textbf{Acc}
         & $\Delta_\textbf{dist}$
         & $\Delta_\textbf{cost}$
         & $\Delta_\textbf{ind}$
         & \textbf{Acc}
         & $\Delta_\textbf{dist}$
         & $\Delta_\textbf{cost}$
         & $\Delta_\textbf{ind}$
         & \textbf{Acc}
         & $\Delta_\textbf{dist}$
         & $\Delta_\textbf{cost}$
         & $\Delta_\textbf{ind}$
         & \textbf{Acc}
         & $\Delta_\textbf{dist}$
         & $\Delta_\textbf{cost}$
         & $\Delta_\textbf{ind}$
         & \textbf{Acc}
         & $\Delta_\textbf{dist}$
         & $\Delta_\textbf{cost}$
         & $\Delta_\textbf{ind}$
         \\
\midrule
    SVM$(\Xb, A)$
        & 86.5 & 0.96 & 0.40 & 1.63
        & 89.5 & 1.18 & 0.44 & 2.11
        & \textbf{88.2} & 0.65 & 0.27 & 2.32
        & 90.8 & 0.05 & \textbf{0.00} & 1.09
        & \textbf{91.1} & 0.07 & 0.03 & 1.06
        & 90.6 & 0.04 & 0.03 & 1.40
        \\
    LR$(\Xb, A)$
        & \textbf{86.7} & 0.48 & 0.50 & 1.91
        & 89.5 & 0.63 & 0.53 & 2.11
        & 87.7 & 0.40 & 0.34 & 2.32
        & 90.5 & 0.08 & 0.03 & 1.06
        & 90.6 & 0.09 & \textbf{0.01} & 1.00
        & 90.6 & 0.19 & 0.22 & 1.28
        \\
    SVM$(\Xb)$
        & 86.4 & 0.99 & 0.42 & 1.80
        & 89.4 & 1.61 & 0.61 & 2.11
        & 88.0 & 0.56 & 0.29 & 2.79
        & \textbf{91.4} & 0.13 & \textbf{0.00} & 0.92
        & 91.0 & 0.17 & 0.08 & 1.09
        & \textbf{91.0} & \textbf{0.02} & 0.03 & 1.64
        \\
    LR$(\Xb)$
        & 86.6 & 0.47 & 0.53 & 1.80
        & 89.5 & 0.64 & 0.57 & 2.11
        & 87.7 & 0.41 & 0.43 & 2.79
        & 91.0 & 0.12 & 0.03 & 1.01
        & 90.6 & 0.13 & 0.10 & 1.65
        & 90.9 & 0.08 & 0.06 & 1.66
        \\ \hline
    FairSVM$(\Xb,A)$
        & 68.1 & \textbf{0.04} & 0.28 & 1.36
        & 66.8 & 0.26 & \textbf{0.12} & 0.78
        & 66.3 & 0.25 & 0.21 & 1.50
        & 90.1 & \textbf{0.02} & \textbf{0.00} & 1.15
        & 90.7 & 0.06 & 0.04 & 1.16
        & 90.3 & 0.37 & \textbf{0.02} & 1.64
        \\ \hline
    SVM$(\Xb_{\nd})$
        & 65.5 & 0.05 & 0.06 & \textbf{0.00}
        & 67.4 & \textbf{0.15} & 0.17 & \textbf{0.00}
        & 65.9 & 0.31 & 0.37 & \textbf{0.00}
        & 66.7 & 0.10 & 0.06 & \textbf{0.00}
        & 58.4 & 0.05 & 0.06 & \textbf{0.00}
        & 62.0 & 0.13 & 0.11 & \textbf{0.00}
        \\
    LR$(\Xb_{\nd})$
        & 65.3 & 0.05 & \textbf{0.05} & \textbf{0.00}
        & 67.3 & 0.18 & 0.18 & \textbf{0.00}
        & 65.6 & 0.31 & 0.31 & \textbf{0.00}
        & 64.7 & \textbf{0.02} & 0.04 & \textbf{0.00}
        & 58.4 & \textbf{0.02} & 0.02 & \textbf{0.00}
        & 61.1 & \textbf{0.02} & 0.03 & \textbf{0.00}
        \\
    SVM$(\Xb_{\nd}, \Ub_{\d})$
        & 86.5 & 0.96 & 0.58 & \textbf{0.00}
        & \textbf{89.6} & 1.07 & 0.70 & \textbf{0.00}
        & 88.0 & \textbf{0.21} & \textbf{0.14} & \textbf{0.00}
        & 90.7 & \textbf{0.02} & 0.03 & \textbf{0.00}
        & \textbf{91.1} & 0.15 & 0.11 & \textbf{0.00}
        & 90.1 & 0.15 & 0.12 & \textbf{0.00}
        \\
    LR$(\Xb_{\nd}, \Ub_{\d})$
        & \textbf{86.7} & 0.43 & 0.90 & \textbf{0.00}
        & 89.5 & 0.35 & 0.77 & \textbf{0.00}
        & 87.8 & 0.14 & 0.34 & \textbf{0.00}
        & 90.9 & 0.28 & 0.05 & \textbf{0.00}
        & 90.9 & 0.49 & 0.07 & \textbf{0.00}
        & 90.2 & 0.43 & 0.21 & \textbf{0.00}
        \\
\bottomrule
    \end{tabular}
    }
    \caption{%
    Results for our numerical simulations from%
    ~\cref{sec:experiments_numerical}, comparing various classifiers differing mostly in their input sets with respect to accuracy (Acc, higher is better) and different recourse fairness metrics ($\Delta_{(\cdot)}$, lower is better) on a number of synthetic datasets (columns).
    SVM: support vector machine, LR: logistic regresion; the first four rows are baselines, the middle row corresponds to the method of~\cite{gupta2019equalizing}, and the last four rows are methods taking causal structure into account.
    For each dataset and metric, the best performing methods are highlighted in \textbf{bold}.
    As can be seen, only our causally-motivated methods (last four rows) achieve  individually fair recourse ($\Delta_{\textbf{ind}}=0$) throughout.
    }
    \label{tab:results}
\end{table*}

\section{Experiments}
\label{sec:experiments}
We perform two sets of experiments.
First, we verify our main claims in numerical simulations~(\cref{sec:experiments_numerical}).
Second, we use our causal measures of fair recourse to conduct a preliminary case study on the Adult dataset~(\cref{sec:experiments_adult}).
We 
refer to Appendix~\ref{sec:app_experimental_details} for further experimental details and to \Cref{sec:app_additional_results} for additional results and analyses.
Code to reproduce our experiments is available at~\href{https://github.com/amirhk/recourse}{https://github.com/amirhk/recourse}.

\subsection{Numerical Simulations}
\label{sec:experiments_numerical}

\paragraph{Data.}
Since computing recourse actions, in general, requires knowledge (or estimation) of the true SCM, we first consider a
controlled setting with two kinds of synthetic data:
\begin{itemize}[]
    \item IMF: the setting underlying IMF recourse where
    features do not causally influence each other, but may depend on the protected attribute~$A$, see~\cref{fig:IMF}. 
    \item CAU:
    features causally depend on each other and on~$A$, see~\cref{fig:CAU}.
    We use $\{X_i:=f_i(A, \PA_i) +U_i\}_{i=1}^n$ with linear (CAU-LIN) and nonlinear (CAU-ANM)~$f_i$.
\end{itemize}
We use $n=3$ non-protected features $X_i$ and a binary protected attribute $A\in\{0,1\}$ in all our experiments
and generate labelled datasets of $N=500$ observations using the SCMs described in more detail in~\Cref{app:scm_specification}.
The ground truth (GT) labels $y^i$ used to train different classifiers are sampled as $Y^i\sim\mathrm{Bernoulli}(h(\xb^i))$ where $h(\xb^i)$ is a linear or nonlinear logistic regression, independently of~$A$, as detailed in~\Cref{app:label_generation}.
\paragraph{Classifiers.}
On each data set, we train several (``fair'') classifiers.
We consider linear and nonlinear logistic regression (LR), and different support vector machines~\citep[SVMs;][]{SchSmo02} (for ease of comparison with~\citet{gupta2019equalizing}), 
trained on 
varying input sets:
\begin{itemize}[]
  \setlength{\itemsep}{0em}
    \item LR/SVM$(\Xb, A)$: trained on all features (\textit{na\"ive baseline});
    \item LR/SVM$(\Xb)$: trained only on non-protected features~$\Xb$ (\textit{unaware baseline});
    \item FairSVM$(\Xb, A)$: the method of~\citet{gupta2019equalizing}, designed to equalise the average distance to the decision boundary across different protected groups;
    \item LR/SVM$(\Xb_{\nd})$: trained only on features $\Xb_{\nd(A)}$ which are non-descendants of $A$, see~\cref{sec:restricted_inputs};
    \item LR/SVM$(\Xb_{\nd}, \Ub_{\d})$: trained on  non-descendants~$\Xb_{\nd(A)}$ of $A$ and on the unobserved variables $\Ub_{\d(A)}$ corresponding to features $\Xb_{\d(A)}$ which are descendants of~$A$, see~\cref{sec:abduction}.
\end{itemize}
To make distances comparable across classifiers, we use either a linear or polynomial kernel for all SVMs (depending on the GT labels) and select all remaining hyperparameters (including the trade-off parameter $\lambda$ for FairSVM) using 5-fold cross validation.
Results for kernel selection by cross-validation are also provided in~\cref{tab:appendix_different_kernel_choices} in~\Cref{sec:app_different_kernel_choices}. Linear (nonlinear, resp.) LR is used when the GT labels are generated using linear (nonlinear, resp.) logistic regression, as detailed in~\Cref{app:label_generation}.
\paragraph{Solving the Causal Recourse Optimisation Problem.}
We treat $A$ and all $U_i$ as non-actionable and all $X_i$ as actionable.
For each negatively predicted individual, we discretise the space of feasible actions, compute the efficacy of each action using a \textit{learned approximate} SCM ($\Mcal_\text{KR}$) (following~\citet{karimi2020imperfect}, see~\Cref{sec:app_different_scm_estimates} for details), and select the least costly valid action resulting in a favourable outcome.
Results using the true oracle SCM ($\Mcal^\star$) and a linear estimate thereof ($\Mcal_\text{LIN}$) are included in~\Cref{tab:appendix_different_SCM_estimates,tab:appendix_different_kernel_choices} in~\Cref{sec:app_different_scm_estimates}; the trends are mostly the same as for $\Mcal_\text{KR}$.

\paragraph{Metrics.}
We report (a) accuracy (\textbf{Acc}) on a held out test set of size 3000; and (b) fairness of recourse as measured by average distance to the boundary ($\Delta_\textbf{dist}$, ~\cref{def:IW-fair})~\cite{gupta2019equalizing}, and our causal group-level  ($\Delta_\textbf{cost}$, ~\cref{def:MINT-fair}) and individual level ($\Delta_\textbf{ind}$, ~\cref{def:counterfactually-MINT-fair}) criteria.
For
(b), we select 50 negatively classified individuals from each protected group and report the difference in group-wise means ($\Delta_\textbf{dist}$ and $\Delta_\textbf{cost}$) or the maximum difference over all 100 individuals ($\Delta_\textbf{ind}$).
To facilitate a comparison between the different SVMs, $\Delta_\textbf{dist}$ is reported in terms of absolute distance to the decision boundary in units of margins.
As a cost function in the causal recourse optimisation problem, we use the
L2 distance between the intervention value $\thetaI$ and the factual value of the intervention targets $\xb_\Ical^\Ftt$.

\paragraph{Results.}
Results are shown in~\Cref{tab:results}.
We find that 
the \textit{na\"ive} and \textit{unaware} baselines
generally exhibit high accuracy and rather poor performance in terms of fairness metrics, but achieve surprisingly low~$\Delta_\textbf{cost}$ on some datasets.
We observe no clear preference of one baseline over the other,
consistent with prior work showing that blindness to protected attributes is not necessarily beneficial for fair \textit{prediction}~\citep{dwork2012fairness%
}; our results suggest this is also true for fair \textit{recourse}. 

FairSVM generally performs well in terms of $\Delta_\textbf{dist}$ (which is what it is trained for), especially on the two IMF datasets, and sometimes (though not consistently) outperforms the baselines on the causal fairness metrics. However, this comes at decreased accuracy, particularly on linearly-separable data.

Both of our causally-motivated setups, LR/SVM$(\Xb_{\nd(A)})$
and LR/SVM$(\Xb_{\nd(A)}, \Ub_{\d(A)})$, achieve $\Delta_\textbf{ind}=0$ throughout \textit{as expected per}~\cref{prop:nondescendants_imply_fair_recourse}, and they are the only methods to do so.
Whereas the former comes at a substantial drop in accuracy due to access to fewer predictive features
(see~\cref{sec:restricted_inputs}), the latter maintains high accuracy by additionally relying on (the true) $\Ub_{\d(A)}$ for prediction.
Its accuracy should be understood as an upper bound on what is possible while preserving ``individually CAU-fair'' recourse if abduction is done correctly, see the discussion in~\cref{sec:abduction}.

Generally, we observe no clear relationship between the different fairness metrics:
e.g., low $\Delta_\textbf{dist}$ does not imply low $\Delta_\textbf{cost}$ (nor vice versa) justifying the need for taking causal relations between features into account (if present) 
to enforce fair recourse at the group-level.
Likewise, \textit{neither small $\Delta_\textbf{dist}$ nor small $\Delta_\textbf{cost}$ imply small $\Delta_\textbf{ind}$, consistent with}~\cref{prop:group_level_insufficient_for_individual_level}, and, empirically, the converse does not hold either.

\paragraph{Summary of Main Findings from~\cref{sec:experiments_numerical}:}
    The non-causal metric~$\Delta_\textbf{dist}$ does not accurately capture recourse unfairness on the CAU-datasets where causal relations are present, thus necessitating our new causal metrics~$\Delta_\textbf{cost}$ and $\Delta_\textbf{ind}$.
    Methods designed in accordance with~\cref{prop:nondescendants_imply_fair_recourse} indeed guarantee individually fair recourse, and group fairness does not imply individual fairness, as expected per~\cref{prop:group_level_insufficient_for_individual_level}.

\subsection{Case Study on the Adult Dataset}
\label{sec:experiments_adult}
\paragraph{Data.}
We use
 the Adult dataset \cite{lichman2013uci}, which consists of $45\text{k}+$ samples without missing data.
We process the dataset similarly to \citet{chiappa2019path} and \citet{nabi2018fair} and adopt the causal graph assumed therein, see~\cref{fig:adult}.
The eight heterogeneous variables include
the three binary protected attributes sex (m=male, f=female), 
age (binarised as $\II\{\text{age} \geq 38\}$; y=young, o=old), and nationality (Nat; US vs non-US),
as well as five non-protected features: marital status (MS; categorical), education level (Edu; integer), working class (WC; categorical), occupation (Occ; categorical), and hours per week (Hrs; integer).
We treat the protected attributes and marital status as non-actionable, and the remaining variables as actionable when searching for recourse actions.

\definecolor{FactualColor}{rgb}{0.85,1,1}
\definecolor{WorstCFColor}{rgb}{1,0.85,1}
\begin{table*}[t]
    \ra{1.5}
    \centering
    \resizebox{\textwidth}{!}{
    \begin{tabular}{lllllllllll}
    \toprule
    Type        & Sex    & Age   & Nat    & MS            & Edu                  & WC                       & Occ     & Hrs & Recourse action & Cost\\
    \midrule
    \texttt{CF} & m  & y & US     & Married   & Some Collg.          & Private                  & Sales   & 32.3 & $do(\{\text{Edu}: \text{Prof-school}, \text{WC}: \text{Private}\})$                                     & 6.2  \\
    \texttt{CF} & m  & y & non-US & Married   & HiSch. Grad          & Private                  & Sales   & 27.8 & $do(\{\text{WC}: \text{Self-empl. ($\lnot$ inc.)}, \text{Hrs}: 92.0\})$                                 & 64.2 \\
    \rowcolor{WorstCFColor}
    \texttt{CF} & m  & o   & US     & Married   & Some Collg. / Bachelors & Private                  & Cleaner & 36.2 & $do(\{\text{Edu}: \text{Prof-school}, \text{WC}: \text{Private}\})$                                     & 5.5  \\
    \texttt{CF} & m  & o   & non-US & Married   & HiSch. Grad          & Private                  & Sales   & 30.3 & $do(\{\text{WC}: \text{Self-empl. ($\lnot$ inc.)}, \text{Hrs}: 92.0\})$                                 & 61.7 \\
    \texttt{CF} & f& y & US     & Married   & Some Collg.          & Self-empl. ($\lnot$inc.) & Sales   & 27.3 & $do(\{\text{Hrs}: 92.0\})$                                                                              & 64.7 \\
    \texttt{CF} & f& y & non-US & Married   & HiSch. Grad          & Self-empl. ($\lnot$inc.) & Sales   & 24.0 & $do(\{\text{Edu}: \text{Some Collg.}, \text{WC}: \text{Self-empl. ($\lnot$ inc.)}, \text{Hrs}: 92.0\})$ & 68.0 \\
    \texttt{CF} & f& o   & US     & Married   & HiSch. / Some Collg. & Private                  & Sales   & 28.8 & $do(\{\text{Edu}: \text{Prof-school}, \text{WC}: \text{Private}\})$                                     & 6.4  \\
    \rowcolor{FactualColor}
    \texttt{F}  & f& o   & non-US & Married   & HiSch. Grad          & W/o pay                  & Sales   & 25   & $do(\{\text{Hrs}: 92.0\})$                                                                              & 67.0 \\
    \bottomrule
    \end{tabular}
    }
     \caption{%
     Individual-level recourse discrimination on the Adult dataset~(\cref{sec:experiments_adult}). Factual (\texttt{F}) observation highlighted in cyan, counterfactual (\texttt{CF}) twin with largest individual-level recourse difference in magenta.
    Consistent with the group-level trends, we observe quantitative discrimination across each protected attribute (favouring older age, male gender, and US nationalism), and qualitative differences in the suggested recourse actions across groups (e.g., favourable predictions based on higher education for men and more working hours for non-US nationals).}
    \label{tab:twins_analysis}
\end{table*}

\paragraph{Experimental Setup.}
We 
extend the probabilistic framework of~\citet{karimi2020imperfect} to 
consider causal recourse in the presence of heterogeneous features, see~\Cref{sec:app_different_scm_estimates} for more details.
We use a
nonlinear LR$(\Xb)$ 
as a classifier (i.e., fairness through unawareness)
 which attains $78.4\%$ accuracy, and (approximately) solve the recourse optimisation problem~\eqref{eq:causal_recourse} 
using brute force search 
as in~\cref{sec:experiments_numerical}. %
We compute the best recourse actions for 10 (uniformly sampled) negatively predicted individuals  from each of the
eight different protected groups (all $2^3$ combinations of the three protected attributes), as well as for each of their seven counterfactual twins, and evaluate using the same metrics as in~\cref{sec:experiments_numerical}.

\paragraph{Results.}
At the group level, we obtain $\Delta_\textbf{dist}=0.89$ and $\Delta_\textbf{cost}=33.32$, indicating group-level recourse discrimination.
Moreover, the maximum difference in \textit{distance} is between \textit{old US males} and \textit{old non-US females} (latter is furthest from the boundary), while that in \textit{cost} is between \textit{old US females} and \textit{old non-US females} (latter is most costly).
This quantitative and qualitative difference between $\Delta_\textbf{dist}$ and $\Delta_\textbf{cost}$ emphasises the general need to account for causal-relations in fair recourse, as present in the Adult dataset.

At the individual-level, we find an average difference in recourse cost to the counterfactual twins of $24.32$ and a maximum difference ($\Delta_\textbf{ind}$) of $61.53$. 
The corresponding individual/factual observation for which this maximum is obtained is summarised along with its seven counterfactual twins in~\cref{tab:twins_analysis}, see the caption for additional analysis.

\paragraph{Summary of Main Findings from~\cref{sec:experiments_adult}:}
Our causal fairness metrics reveal qualitative and quantitative aspects of recourse discrimination  at both the group and individual level.
In spite of efforts to design classifiers that are predictively fair, recourse unfairness remains a valid concern on real datasets. 

\section{On Societal Interventions}
\label{sec:social_interventions}
Our notions of fair causal recourse~(\cref{def:MINT-fair,def:counterfactually-MINT-fair}) depend on multiple components $(\Dcal, h, c, \Mcal)$.
As discussed in~\cref{sec:introduction}, 
in fair ML, the typical procedure is to \textit{alter the classifier} $h$.
This is the approach proposed for Equalizing Recourse by~\citet{gupta2019equalizing}, which we have discussed in the context of fair \textit{causal} recourse~(\cref{sec:achieving_recourse}) and explored experimentally~(\cref{sec:experiments}).
However, requiring the learnt classifier~$h$ to satisfy some constraint implicitly places the cost of an intervention on the deployer. 
For example, a bank might need to modify their classifier so as to offer credit cards to some individuals who would not otherwise receive them. 

Another possibility is to \textit{alter the data-generating process}
(as captured by the SCM~$\Mcal$ and manifested in the form of the observed data~$\Dcal$)
via a \textit{societal intervention} 
in order to achieve fair causal recourse
with a \textit{fixed} classifier~$h$.
By considering changes to the underlying SCM or to some of its mechanisms, we may facilitate outcomes which are more societally fair overall, and perhaps end up with a dataset that is more amenable to fair causal recourse (either at the group or individual level). 
Unlike the setup of \citet{gupta2019equalizing}, our causal approach here is perhaps particularly well suited to exploring this perspective, as we are already explicitly modelling the causal generative process, i.e., how changes to parts of the system will affect the other variables.
We demonstrate our ideas for the toy example
with different variances across groups from~\Cref{fig:different_variances}.
Here, the difference in recourse cost across groups cannot easily be resolved by changing the classifier~$h$ (e.g., per the techniques in~\cref{sec:achieving_recourse}):
to achieve perfectly fair recourse, we would have to use a constant classifier, i.e., either approve all credit cards, or none, irrespective of income.
Essentially, changing $h$ does not address the root of the problem, namely the discrepancy in the two populations.
Instead, we investigate how to reduce the larger cost of recourse within the higher-variance group by altering the data generating process via societal interventions.

\begin{figure*}[t]
    \begin{subfigure}[b]{0.5\textwidth}
        \centering
        \includegraphics[width=.85\columnwidth]{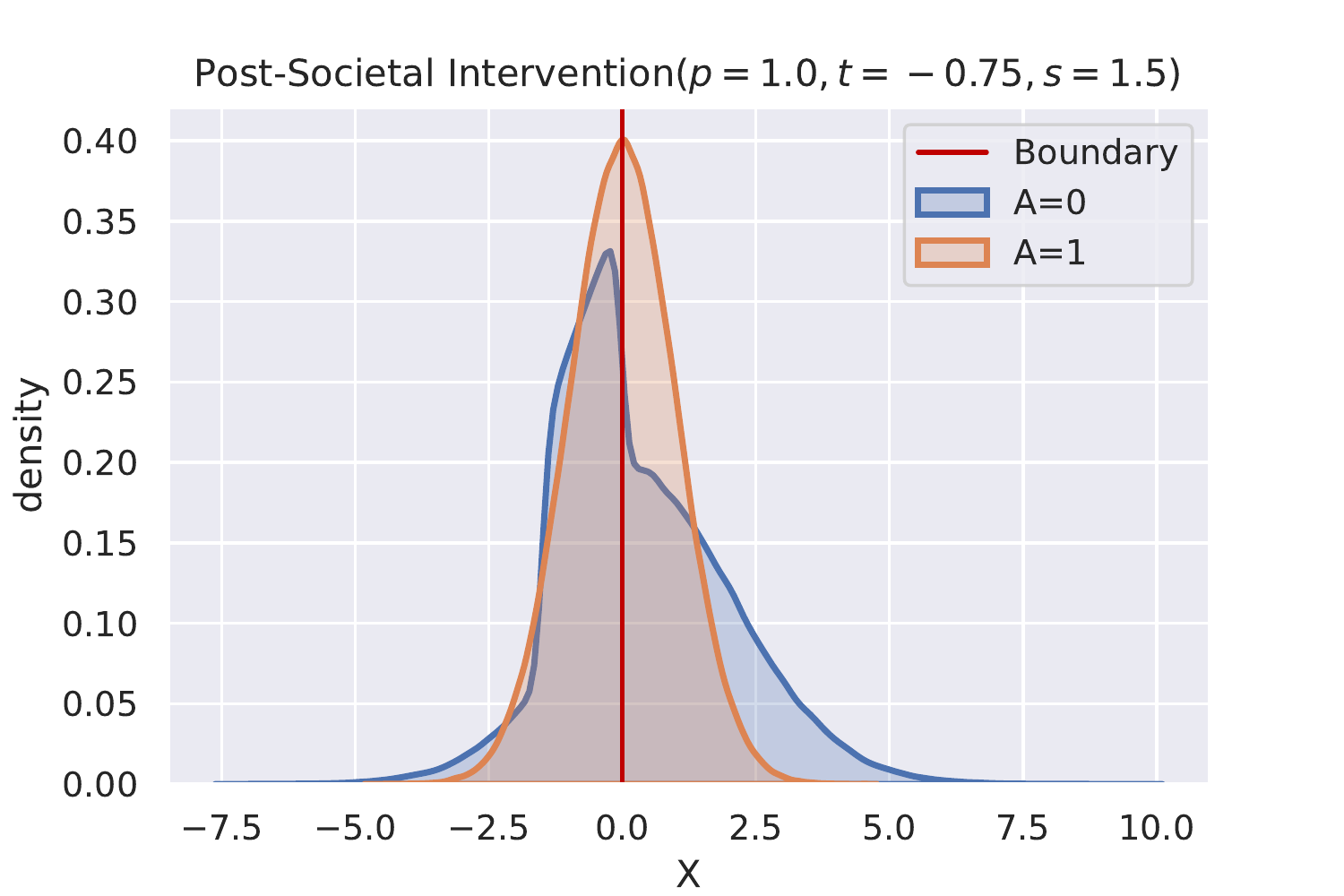}
        \caption{Post-intervention distribution}
        \label{fig:SI_dist}
    \end{subfigure}%
    \begin{subfigure}[b]{0.5\textwidth}
        \centering
        \includegraphics[width=.85\columnwidth]{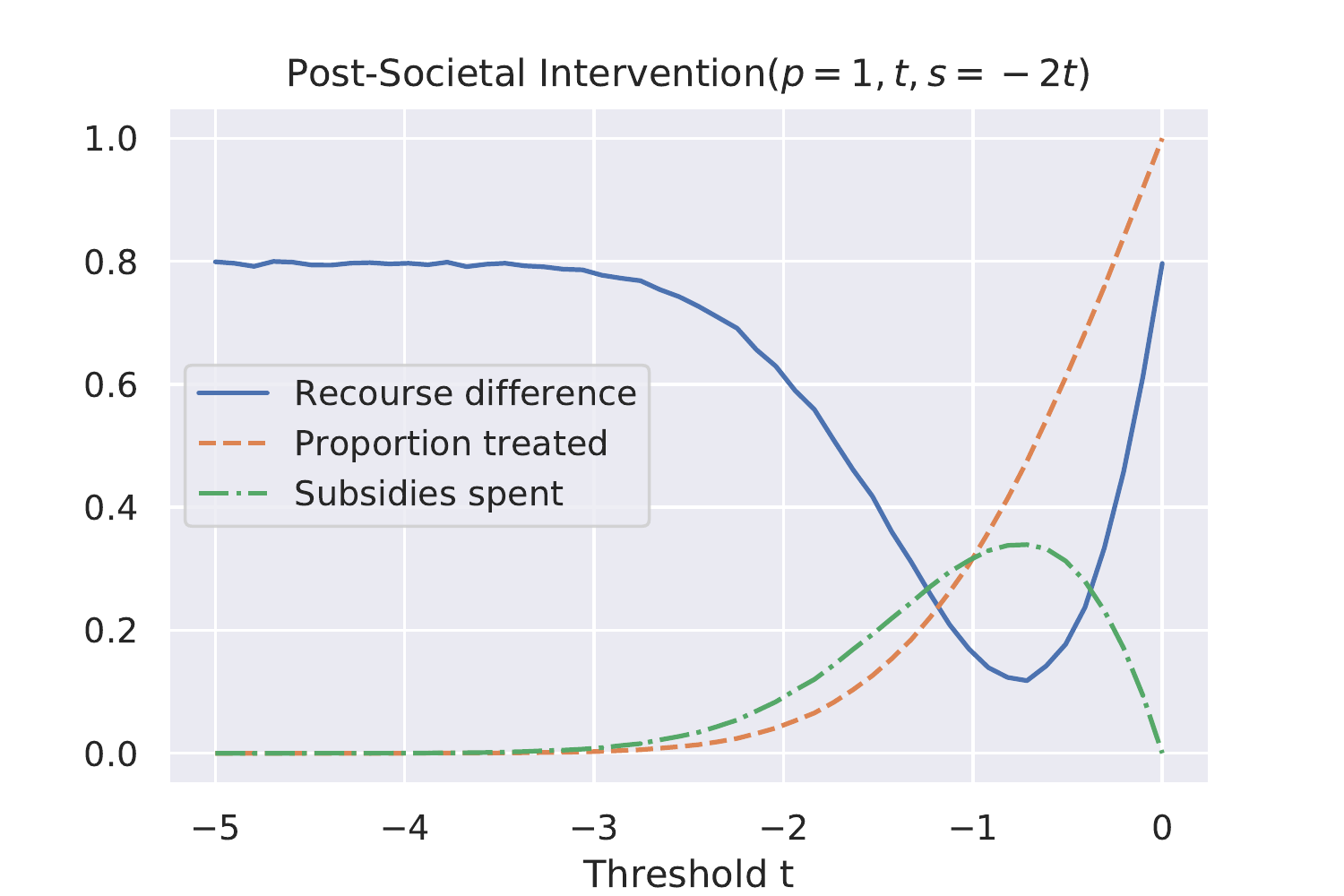}
        \caption{Comparison of different societal interventions}
        \label{fig:SI_costs}
    \end{subfigure}
    \caption{%
    \text{(a)} Distribution after applying a societal intervention to the credit-card example from~\Cref{fig:different_variances}.
    We randomly select a \textit{proportion} $p=1$ of individuals from the disadvantaged group (blue, $A=0$) to receive a \textit{subsidy} $s=1.5$ if $U_X$ is below the \textit{threshold} $t=-0.75$.
    As a result, the distribution of negatively-classified individuals ($X<0$) shifts towards the boundary which makes it more similar to those in $A=1$, thus resulting in fairer recourse. At the same time, the distribution of positively-classified individuals ($X>0$) remains unchanged.
    \text{(b)}
    Comparison of different societal interventions $i_k=(1, t, -2t)$ with respect to their benefit (reduction in recourse difference) and cost (paid-out subsidies). The threshold $t\approx -0.75$ (corresponding to the distribution shown on the left) leads to the largest reduction in recourse difference, but also incurs the highest cost.
    Smaller reductions can be achieved using two different thresholds: one corresponding to giving a larger subsidy to fewer individuals, and the other to giving a smaller subsidy to more individuals.
    }
    \label{fig:post_intervention}
\end{figure*}

Let 
$i_k$ denote a societal intervention that modifies the data generating process, $X:=(2-A)U_X$, $U_X\sim\Ncal(0,1)$, by changing the original SCM $\Mcal$ to $\Mcal'_k=i_k(\Mcal)$.
For example, $i_k$ may introduce additional variables or modify a subset of the original structural equations.
Specifically,  we consider subsidies to particular eligible individuals.
We introduce a new treatment variable $T$ which randomly selects a proportion $0\leq p\leq 1$ of individuals from $A=0$ who are awarded a subsidy $s$ if their latent variable $U_X$ is below a threshold $t$.\footnote{E.g., for interventions with minimum quantum size and a fixed budget, it makes sense to spread  interventions across a \textit{randomly} chosen subset since it is not possible to give everyone a very small amount, see 
~\cite{grgic2017fairness} for broader comments on the potential benefits of randomness in fairness.
Note %
that $p=1$, i.e., deterministic interventions are included as a special case.
}
This is captured by the modified structural equations%
\begin{align*}
\textstyle
    T&:=(1-A)\II\{U_T<p\}, &U_T&\sim \text{Uniform}[0,1], \\
    X&:=(2-A)U_X +sT\II\{U_X < t\}, &U_X&\sim \Ncal(0,1).
\end{align*}

Here, each societal intervention $i_k$ thus corresponds to a particular way of setting the triple $(p, t, s)$.
To avoid changing the predictions $\mathrm{sgn}(X)$, we only consider $t\leq 0$ and $s\leq-2t$.
The modified distribution resulting from $i_k=(1,-0.75,1.5)$ is shown in~\Cref{fig:SI_dist}, see the caption for details. 

To evaluate the effectiveness of different societal interventions $i_k$ in reducing recourse unfairness, we compare their associated societal costs $c_k$ and benefits $b_k$.
Here, the cost $c_k$ of implementing $i_k$ can reasonably be chosen as the total amount of paid-out subsidies, and the benefit $b_k$, as the reduction in the difference of average recourse cost across groups.
We then reason about different societal interventions $i_k$ by  simulating the proposed change via sampling data from $\Mcal'_k$ and computing $b_k$ and $c_k$ based on the simulated data.
To decide which intervention to implement, we compare the societal benefit $b_k$ and cost $c_k$ of $i_k$ for different $k$ and choose the one with the most favourable trade-off.
We show the societal benefit and cost tradeoff for $i_k=(1, t, -2t)$ with varying $t$ in~\cref{fig:SI_costs} and refer to the caption for further details.
Plots similar to~\cref{fig:post_intervention} for different choices of $(p,t,s)$ are shown in~\cref{fig:more_societal_interventions} in~\Cref{sec:app_additional_societal_interventions}.
Effectively, our societal intervention does not change the  outcome of credit card approval but ensures that the effort required (additional income needed) for rejected individuals from two groups is the same.
Instead of using a threshold to select eligible individuals as in the toy example above, for more complex settings, our individual-level unfairness metric (\Cref{def:counterfactually-MINT-fair}) may provide a useful way to inform whom to target with societal interventions as it can be used to identify individuals for whom the counterfactual difference in recourse cost is particularly high.

\section{Discussion}
\label{sec:conclusion}

With data-driven decision systems pervading our societies, establishing appropriate fairness metrics and paths to recourse
are gaining major significance. There is still much work to do in identifying and conceptually understanding the best path forward. Here we make progress towards this goal by applying tools of graphical causality. We are hopeful that this approach will continue to be fruitful  %
as we search together with stakeholders and broader society for the right concepts and definitions, as well as for assaying interventions on societal mechanisms. 

While our fairness criteria may help assess the fairness of recourse, %
it is still unclear how best to achieve fair causal recourse algorithmically.
Here, we argue that fairness considerations may benefit from considering the larger system at play---instead of focusing solely on the classifier---and that a causal model of the underlying data generating process provides a principled framework for addressing issues such as multiple sources of unfairness, as well as different costs and benefits to individuals,  institutions, and society.

Societal interventions to overcome (algorithmic) discrimination constitute a complex topic which not only applies to fair recourse but also to other notions of fairness. It deserves further study well beyond the scope of the present work.
We may also question whether it is %
appropriate to perform a societal intervention on all individuals in a subgroup. 
For example, when considering who is approved for a credit card, an individual might not be able to pay their statements on time and this could imply costs to them, to the bank, or to society.
This idea relates to  the economics literature which studies the effect of policy interventions on society, institutions, and individuals~\citep{heckman2005structural,heckman2010building}. Thus, future work could focus on formalising the effect of these interventions to the \SCM, as such a framework would help trade off the costs and benefits for individuals, companies, and society.

\section*{Acknowledgements}
We are grateful to Chris Russell for insightful feedback on connections to existing fairness notions within machine learning and philosophy.
We also thank Matth\"aus Kleindessner, Adrián Javaloy Bornás, and the anonymous reviewers 
for helpful comments and suggestions.

AHK is appreciative of NSERC, CLS, and Google for generous funding support. UB acknowledges support from DeepMind and
the Leverhulme Trust via the Leverhulme Centre for the Future of Intelligence (CFI), and from the Mozilla Foundation.
AW acknowledges support from a Turing AI Fellowship under grant EP/V025379/1, The Alan Turing Institute under
EPSRC grant EP/N510129/1 \& TU/B/000074, and the Leverhulme Trust via CFI.
This work was supported by the German Federal Ministry of Education and Research (BMBF): Tübingen AI Center, FKZ: 01IS18039B, and by the Machine Learning Cluster of Excellence, EXC number 2064/1 – Project number 390727645.

\bibliography{references}

\clearpage
\appendix

\section{Experimental Details}
\label{sec:app_experimental_details}
In this Appendix, we provide additional details on our experiment setup.

\subsection{SCM Specification}
\label{app:scm_specification}
First, we give the exact form of SCMs used to generate our three synthetic data sets IMF, CAU-LIN, and CAU-ANM.
Besides the desired characteristics of independently-manipulable (IMF) or causally dependent (CAU) features and linear (LIN) or nonlinear (ANM) relationships with additive noise, we choose the particular form of structural equations for each setting such that all features are roughly standardised, i.e., such that they all approximately have a mean of zero and a variance one.

We use the causal structures shown in~\Cref{fig:synthetic_data_graph}. Apart from the desire two make the causal graphs similar to facilitate a better comparison and avoid introducing further nuisance factors while respecting the different structural constraints of the IMF and CAU settings, this particular choice is motivated by having at least one feature which is not a descendant of the protected attribute $A$. This is so that LR/SVM($\Xb_{\nd(A)}$) and LR/SVM($\Xb_{\nd(A)}, \Ub_{\d(A)}$) always have access to at least one actionable variable ($X_2$) which can be manipulated to achieve recourse.

\subsubsection{IMF}
For the IMF data sets, we sample the protected attribute $A$ and the features $X_i$ according to the following SCM:
\begin{align*}
    A &:= 2U_A-1, &U_A\sim \mathrm{Bernoulli}(0.5)\\
    X_1 &:= 0.5A + U_1, &U_1\sim \Ncal(0,1)\\
    X_2 &:= U_2, &U_2\sim\Ncal(0,1)\\
    X_3 &:= 0.5A + U_3, &U_3\sim\Ncal(0,1)
\end{align*}

\subsubsection{CAU-LIN}
For the CAU-LIN data sets, we sample  $A$ and  $X_i$ according to the following SCM:
\begin{align*}
    A &:= 2U_A-1, &U_A\sim \mathrm{Bernoulli}(0.5)\\
    X_1 &:= 0.5A + U_1, &U_1\sim \Ncal(0,1)\\
    X_2 &:= U_2, &U_2\sim\Ncal(0,1)\\
    X_3 &:= 0.5(A+X_1-X_2) + U_3, &U_3\sim\Ncal(0,1)
\end{align*}

\subsubsection{CAU-ANM}
For the CAU-ANM data sets, we sample  $A$ and $X_i$ according to the following SCM:
\begin{align*}
    A &:= 2U_A-1, &U_A\sim \mathrm{Bernoulli}(0.5)\\
    X_1 &:= 0.5A + U_1, &U_1\sim \Ncal(0,1)\\
    X_2 &:= U_2, &U_2\sim\Ncal(0,1)\\
    X_3 &:= 0.5A+0.1(X_1^3 - X_2^3) + U_3, &U_3\sim\Ncal(0,1)
\end{align*}

\subsection{Label generation}
\label{app:label_generation}
To generate ground truth labels on which the different classifiers are trained, we consider both a linear and a nonlinear logistic regression.
Specifically, we generate ground truth labels according to
\begin{equation*}
    Y:=\II\{U_Y<h(X_1,X_2,X_3)\}, \quad \quad U_Y\sim \mathrm{Uniform}[0,1].
\end{equation*}
In the linear case, $h(X_1,X_2,X_3)$ is given by
\begin{equation*}
    h(X_1,X_2,X_3)=\left(1+e^{-2(X_1-X_2+X_3)}\right)^{-1}.
\end{equation*}
In the nonlinear case, $h(X_1,X_2,X_3)$ is given by
\begin{equation*}
    h(X_1,X_2,X_3)=\left(1+e^{4-(X_1+2X_2+X_3)^2}\right)^{-1}.
\end{equation*}

\subsection{Fair model architectures and training hyper-parameters}
We use the implementation of~\citet{gupta2019equalizing} for the FairSVM and the \texttt{sklearn SVC} class~\citep{pedregosa2011scikit} for all other SVM variants.
We consider the following values of hyperparameters (which are the same as those reported in~\cite{gupta2019equalizing} for ease of comparison) and choose the best by 5-fold cross validation (unless stated otherwise):
$\text{kernel type}\in\{\text{linear, poly, rbf}\}$,
regularisation strength $C\in\{1, 10, 100\}$, 
RBF kernel bandwidth $\gamma_\textsc{rbf}\in\{0.001, 0.01, 0.1, 1\}$, polynomial kernel $\text{degree}\in \{2, 3, 5\}$;
following~\cite{gupta2019equalizing}, we also pick the fairness trade-off parameter
$\lambda = \{0.2, 0.5, 1, 2, 10, 50, 100\}$ by cross-validation.

For the nonlinear logistic regression model, we opted for an instance of the \texttt{sklearn MLPClassifier} class with two hidden layers (10 neurons each) and ReLU activation functions. This model was then optimised on its inputs using the default optimiser and training hyperparameters. %

\subsection{Optimisation approach}
Since an algorithmic contribution for solving the causal recourse optimisation problem is not the main focus of this work, we choose to discretise the space of possible recourse actions and select the best (i.e., lowest cost) valid action by performing a brute-force search.
For an alternative gradient-based approach to solving the causal recourse optimisation problem, we refer to~\cite{karimi2020imperfect}.

For each actionable feature $X_i$, denote by $\max_i$ and $\min_i$ its maximum and minimum attained in the training set, respectively. Given a factual observation $x^\Ftt_i$ of $X_i$, we discretise the search space and pick possible intervention values $\theta_i$ using 15 equally-spaced bins in the range $[x^\Ftt_i-2(x^\Ftt_i-\min_i), x^\Ftt_i+2(\max_i-x^\Ftt_i)]$.
We then consider all possible combinations of intervention values over all subsets $\Ical$ of the actionable variables.
We note that for LR/SVM$(\Xb_{\nd})$ and LR/SVM$(\Xb_{\nd}, \Ub_{\d})$, only $X_2$ is actionable, while for the other LR/SVMs all of $\{X_1, X_2, X_3\}$ are actionable.

\subsection{Adult dataset case study}
The causal graph for the Adult dataset informed by expert knowledge~\cite{nabi2018fair,chiappa2019path} is depicted in~\Cref{fig:adult}.

Because the true structural equations are not known, we learn an approximate SCM for the Adult dataset by fitting each parent-child relationship in the causal graph.
Since most variables in the Adult dataset are categorical, additive noise is not an appropriate assumption for most of them.
We therefore opt for modelling each structural equation, $X_i:=f_i(\PA_i,U_i)$,
using a latent variable model; specifically, we use a conditional variational autoencoder (CVAE)~\cite{sohn2015learning}, similar to~\cite{karimi2020imperfect}.

We use deterministic conditional decoders $\dec_i(\PA_i, U_i; \psi_i)$, implemented as neural nets parametrised by $\psi_i$, and use an isotropic Gaussian prior, $U_i\sim\Ncal(0,\Ib)$, for each $X_i$.

For continuous features, the decoders directly output the value assigned to $X_i$, i.e., we approximate the structural equations as
\begin{equation}
    X_i^{\text{continuous}}:= \dec_i(\PA_i, U_i; \psi_i), \quad \quad U_i\sim\Ncal(0,\Ib).
\end{equation}
For categorical features, the decoders output a vector of class probabilities (by applying a softmax operation after the last layer). The arg max is then assigned as the value of the corresponding categorical feature, i.e.,
\begin{equation}
    X_i^{\text{categorical}}:= \argmax \dec_i(\PA_i, U_i; \psi_i), \quad \quad U_i\sim\Ncal(0,\Ib).
\end{equation}
The decoders $\dec_i(\PA_i, U_i; \psi_i)$ are trained using the standard variational framework~\cite{kingma2013auto,rezende2014stochastic}, amortised with approximate Gaussian posteriors $q_{\phi_i}(U_i | X_i,  \PA_i)$ whose means and variances are computed by encoders in the form of neural nets with parameters $\phi_i$. 
For continuous features, we use
the standard reconstruction error between real-valued predictions and targets, i.e., L$2$/MSE$(X_i, \dec_i(\PA_i, U_i; \psi_i))$, whereas for categorical features, we instead use the cross entropy loss between the one-hot encoded value of $X_i$ and the predicted vector of class probabilities, i.e., $\text{CrossEnt}(X_i, \text{softmax}(\dec_i(\PA_i, U_i; \psi_i))$. %

The CVAEs are trained on $6,000$ training samples using a fixed learning rate of $0.05$, and a batch size of $128$ for $100$ epochs with early stopping on a held-out validation set of $250$ samples.
For each parent-child relation, we train 10 models with the number of hidden layers and units randomly drawn from the following configurations for the encoder: $\text{enc}_\text{arch} = \{(\zeta, 2, 2), (\zeta, 3, 3), (\zeta, 5, 5), (\zeta, 32, 32, 32)\}$, where $\zeta$ is the input dimensionality; and similarly for the decoders from: $\text{dec}_\text{arch} = \{(2, \eta), (2, 2, \eta), (3, 3, \eta), (5, 5, \eta), (32, 32, 32, \eta)\}$, where $\eta$ is either one for continuous variables, or alternatively the size of the one-hot embedding for categorical variables (e.g., Work Class,  Marital Status, and Occupation have 7, 10, and 14 categories, respectively).
Moreover, we also randomly pick a latent  dimension from $\{1,3,5\}$.
We then select the model with the smallest MMD score~\cite{gretton2012kernel} between true instances and samples from the decoder post-training.

To perform abduction for counterfactual reasoning with such an approximate CVAE-SCM, we sample $U_i$ from the approximate posterior. For further discussion, we refer to~\cite{karimi2020imperfect}, Appendix~C.

Finally, using this approximate SCM, we solve the recourse optimisation problem similar to the synthetic experiments above.
The caveat with this approach (and any real-world dataset absent a true SCM for that matter) is that we are not able to certify that a given recourse action generated under the assumption of an approximate SCM will guarantee recourse when executed in the true world governed by the real (unknown) SCM.

\section{Additional Results}
\label{sec:app_additional_results}
In this Appendix, we provide additional experimental results omitted from the main paper due to space constraints.

\subsection{Additional Societal Interventions}
\label{sec:app_additional_societal_interventions}
\begin{figure*}[p]
    \centering
    \begin{subfigure}{0.33\textwidth}
        \includegraphics[width=\textwidth]{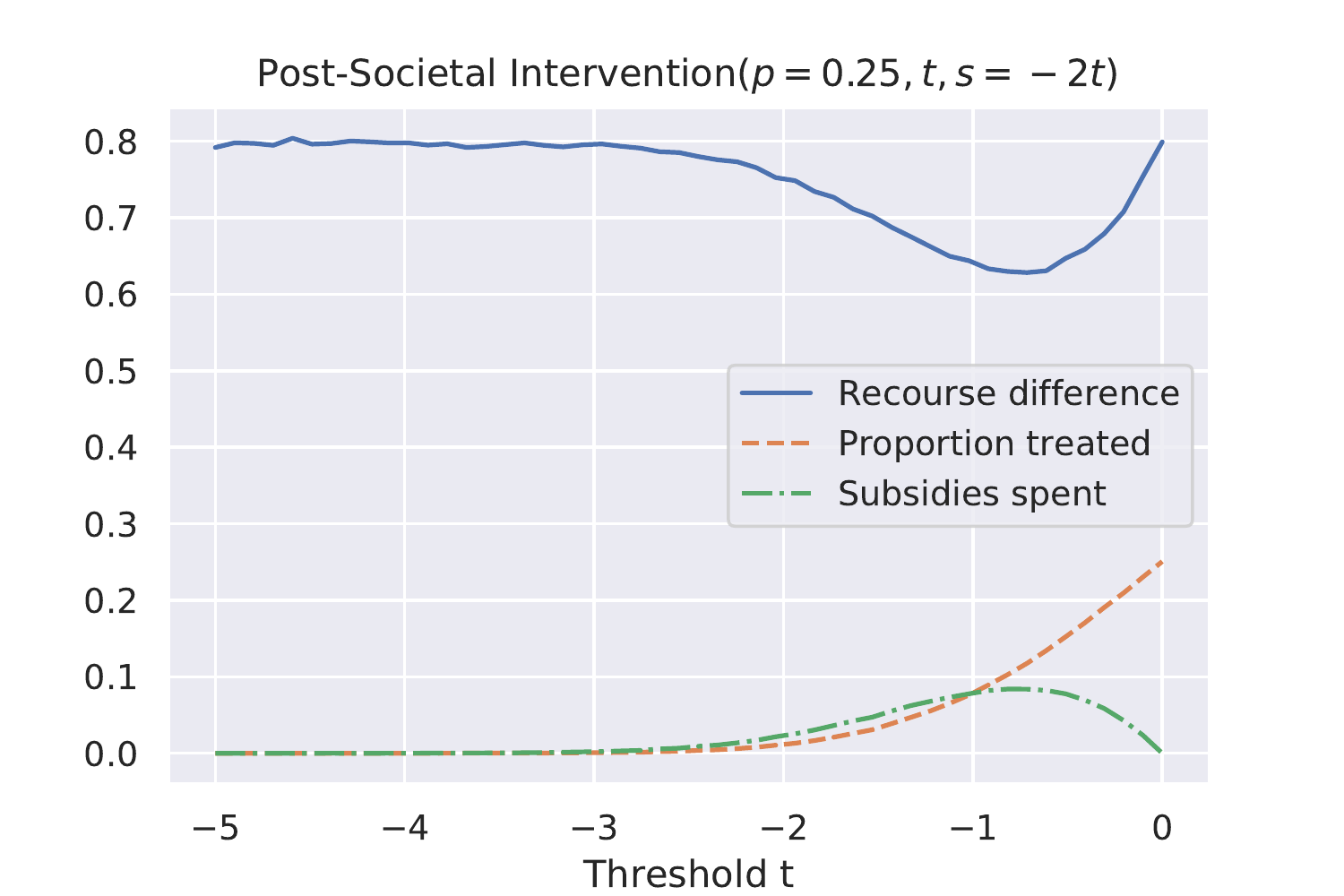}
    \end{subfigure}%
    \begin{subfigure}{0.33\textwidth}
        \includegraphics[width=\textwidth]{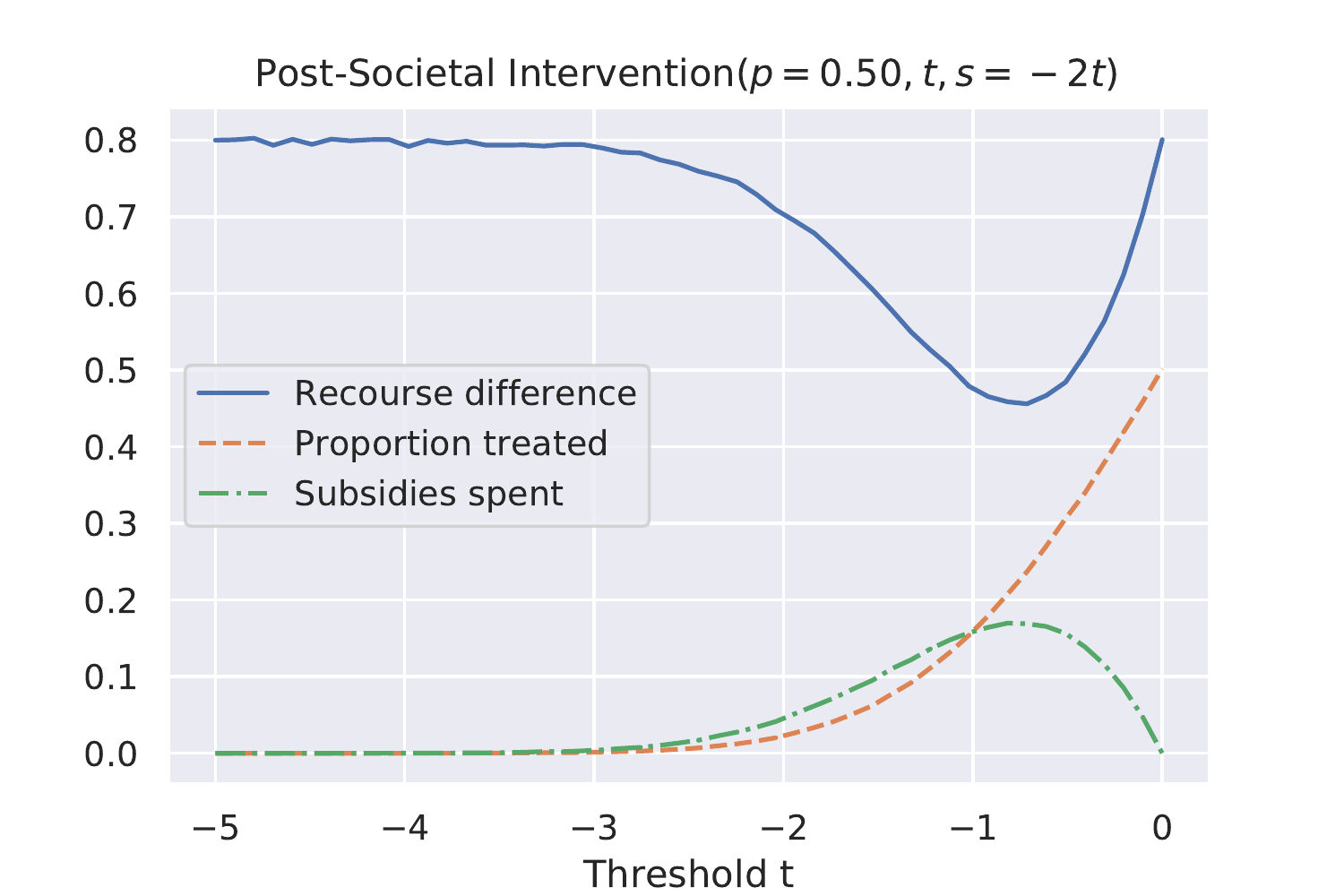}
    \end{subfigure}%
    \begin{subfigure}{0.33\textwidth}
        \includegraphics[width=\textwidth]{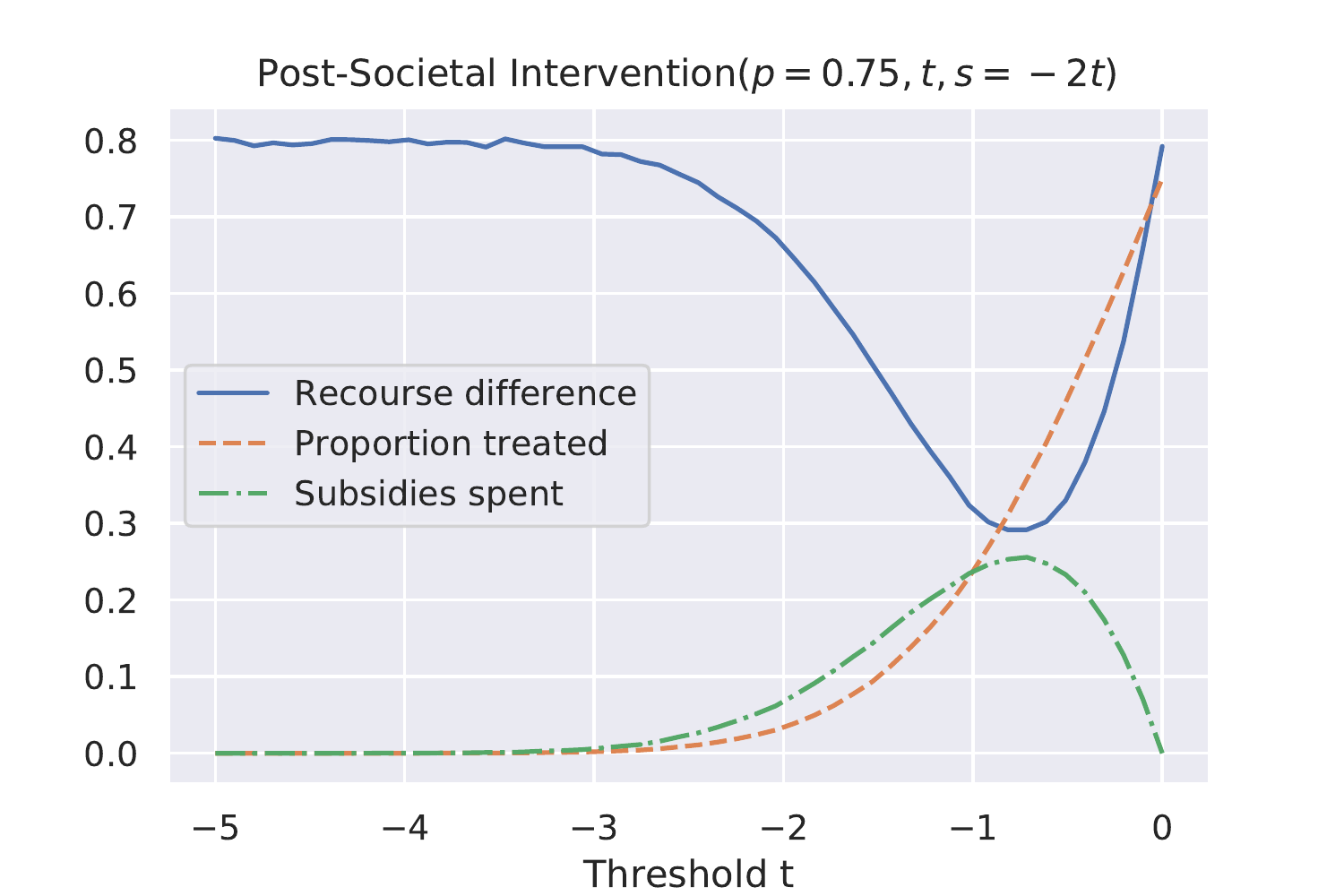}
    \end{subfigure}%
    
    \begin{subfigure}{0.33\textwidth}
        \includegraphics[width=\textwidth]{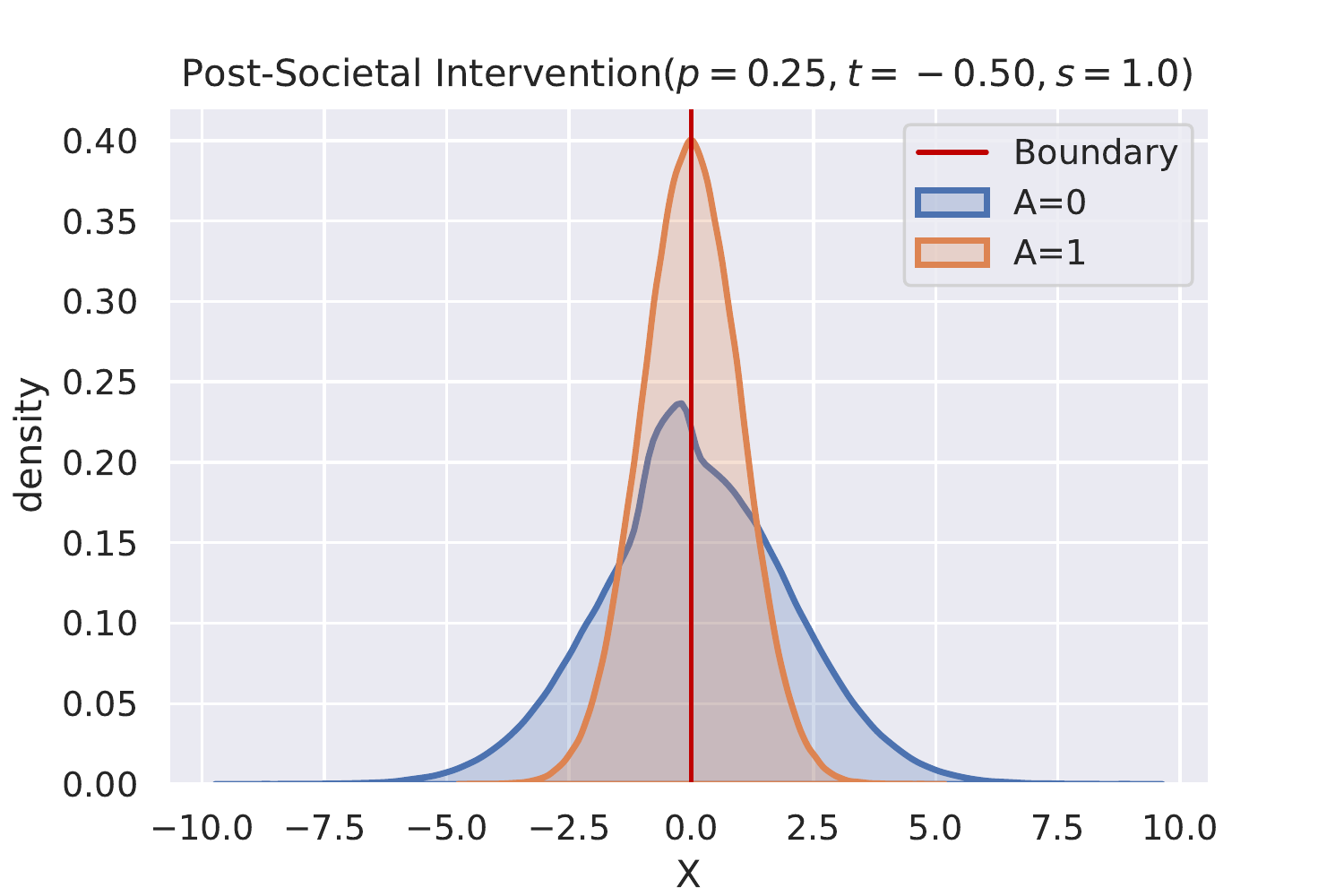}
    \end{subfigure}%
    \begin{subfigure}{0.33\textwidth}
        \includegraphics[width=\textwidth]{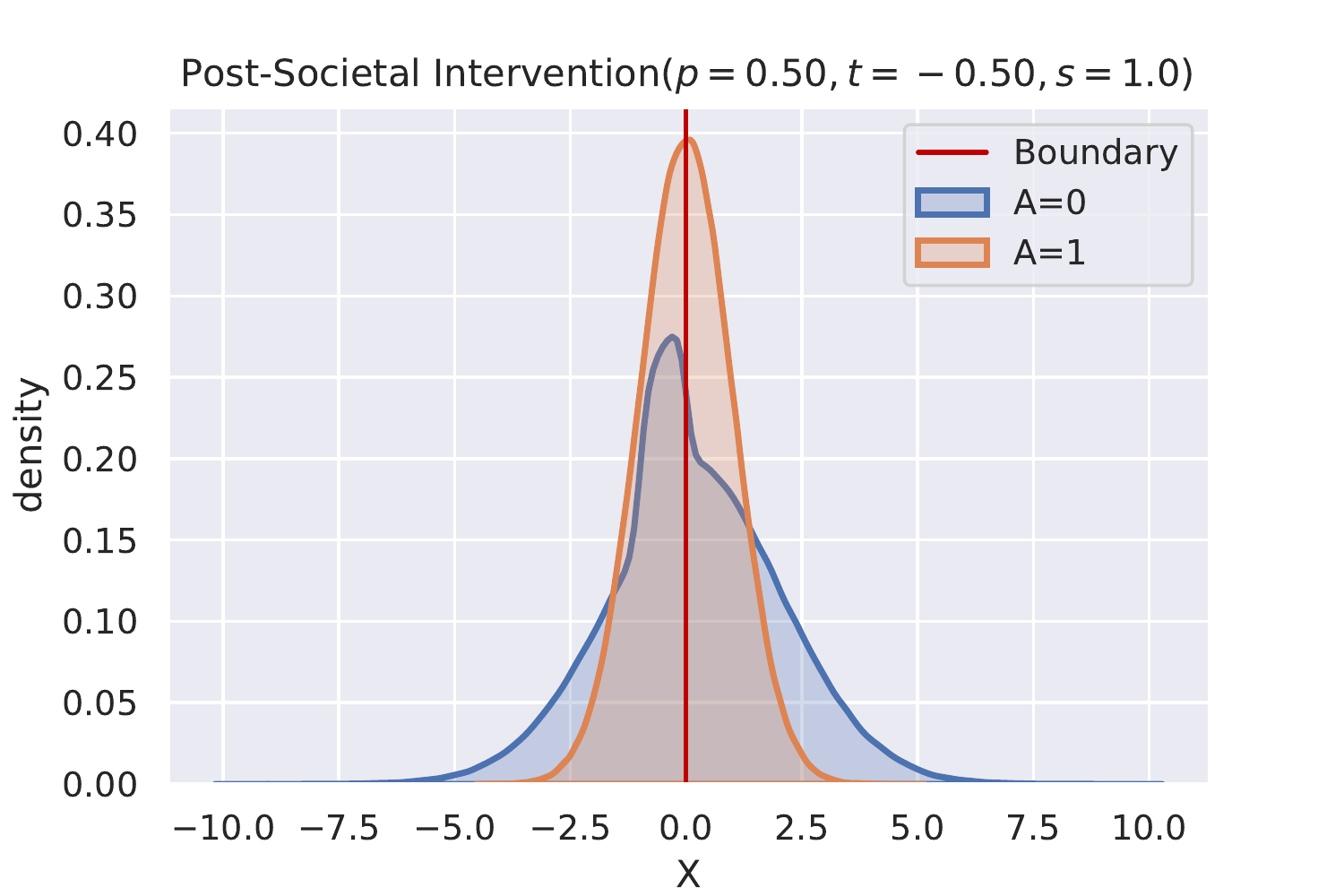}
    \end{subfigure}%
    \begin{subfigure}{0.33\textwidth}
        \includegraphics[width=\textwidth]{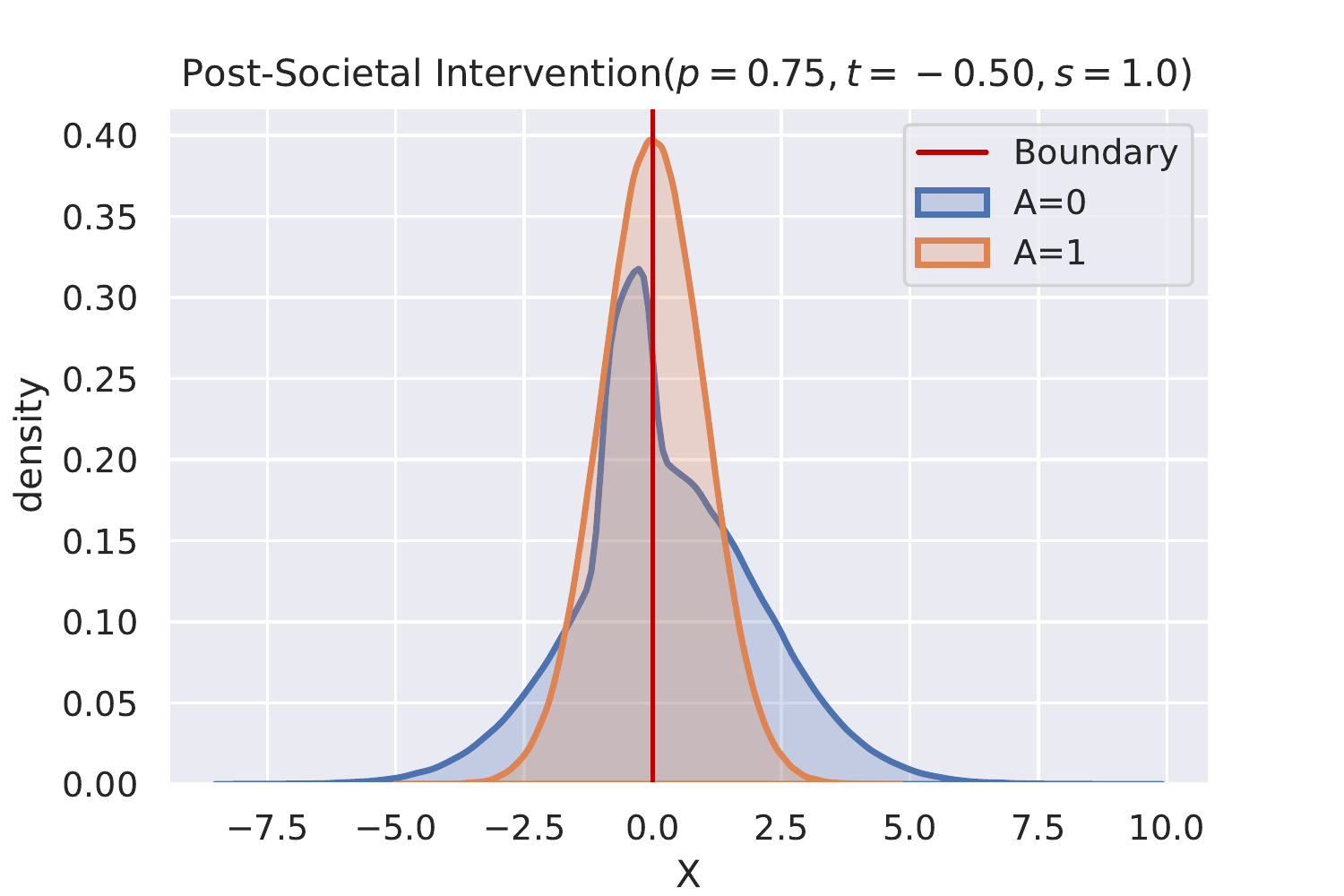}
    \end{subfigure}%
    
    \begin{subfigure}{0.33\textwidth}
        \includegraphics[width=\textwidth]{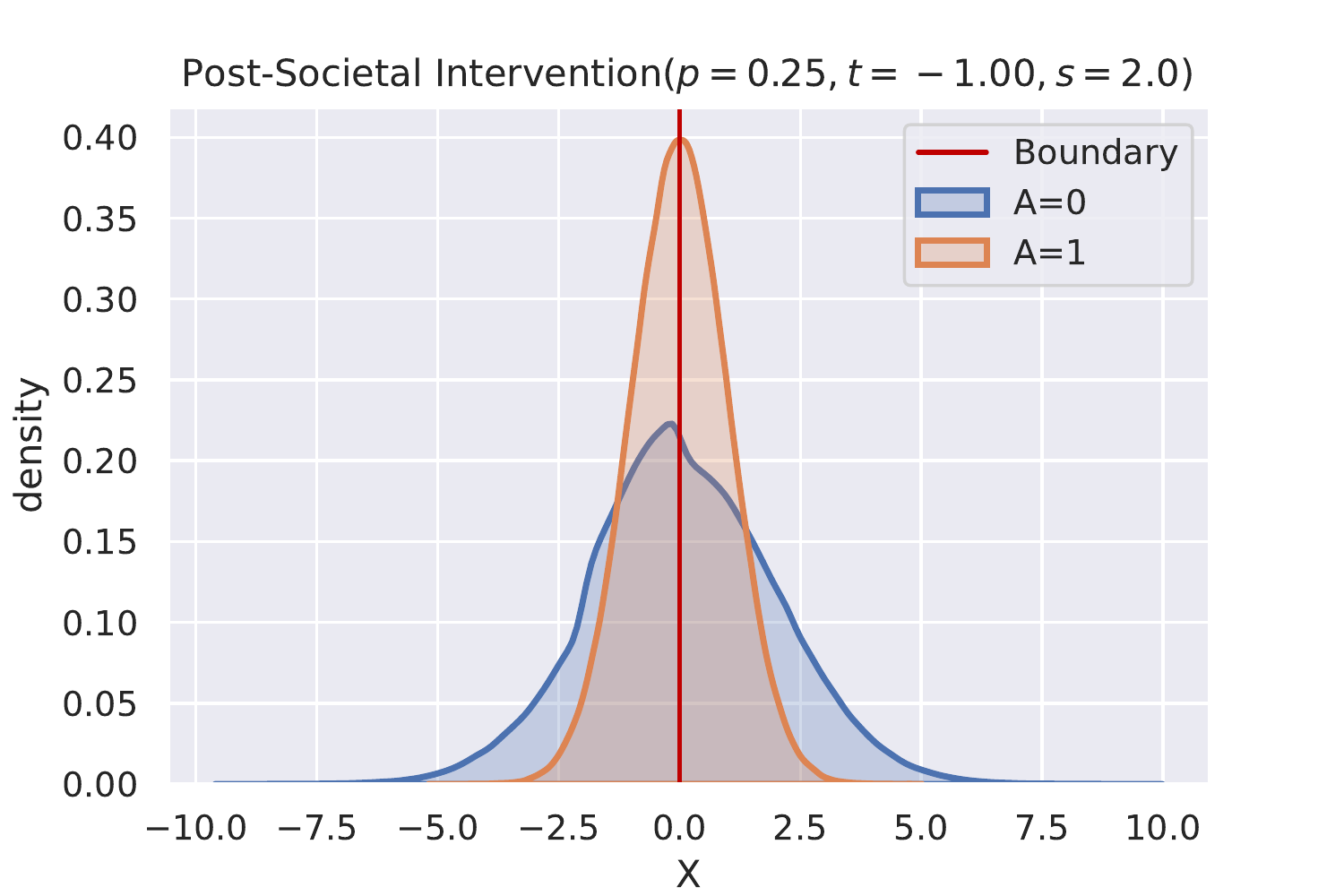}
    \end{subfigure}%
    \begin{subfigure}{0.33\textwidth}
        \includegraphics[width=\textwidth]{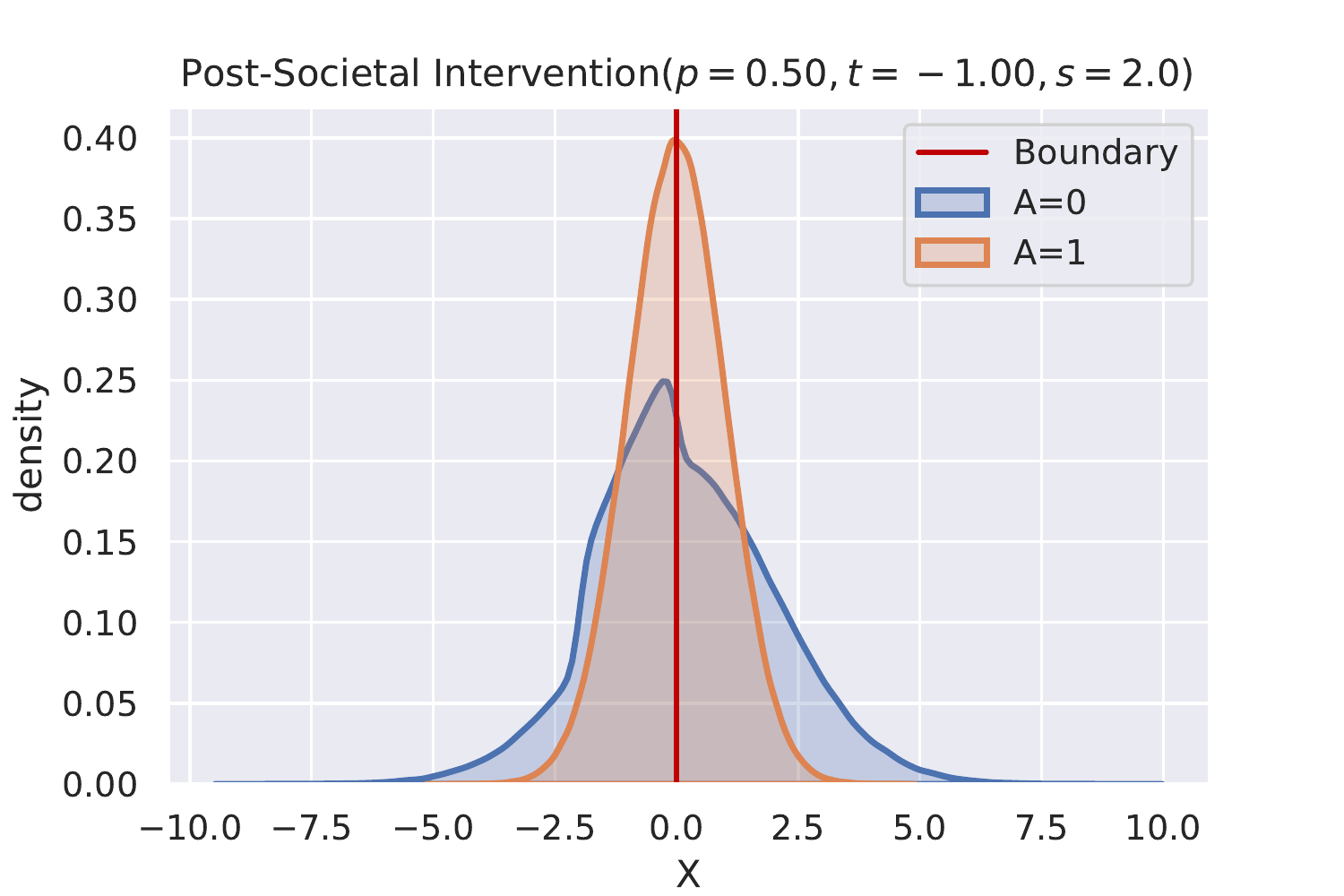}
    \end{subfigure}%
    \begin{subfigure}{0.33\textwidth}
        \includegraphics[width=\textwidth]{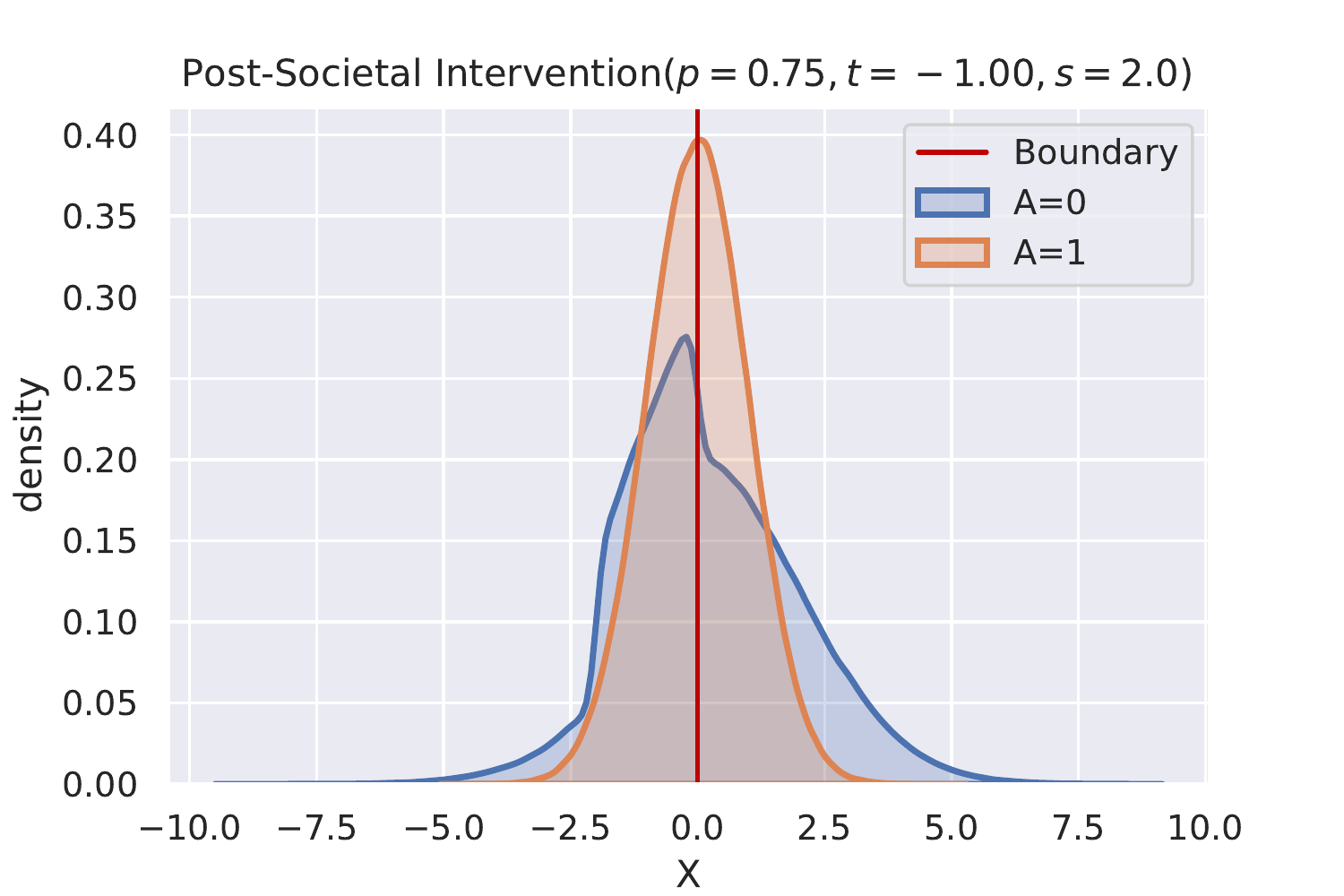}
    \end{subfigure}%
    
    \begin{subfigure}{0.33\textwidth}
        \includegraphics[width=\textwidth]{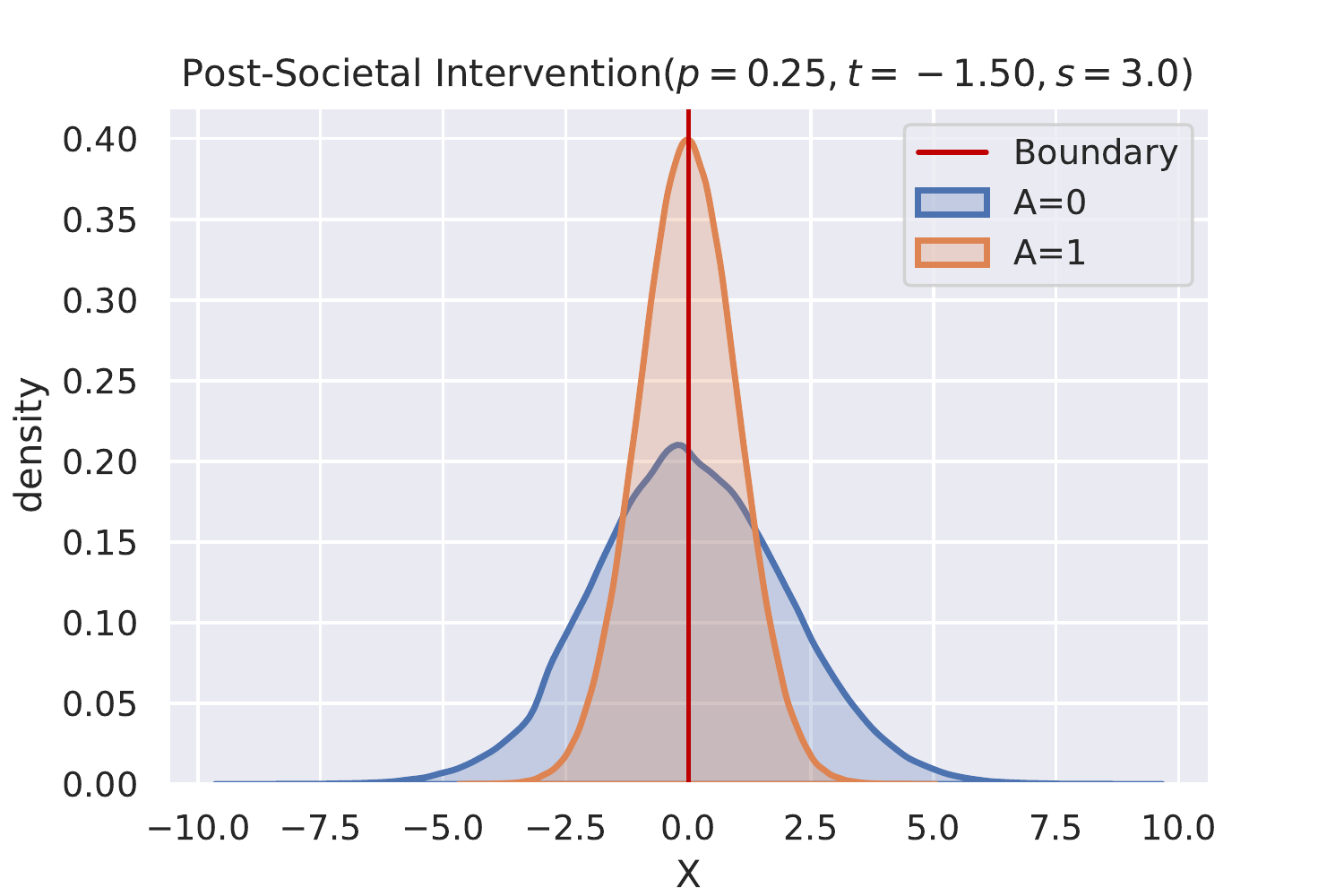}
    \end{subfigure}%
    \begin{subfigure}{0.33\textwidth}
        \includegraphics[width=\textwidth]{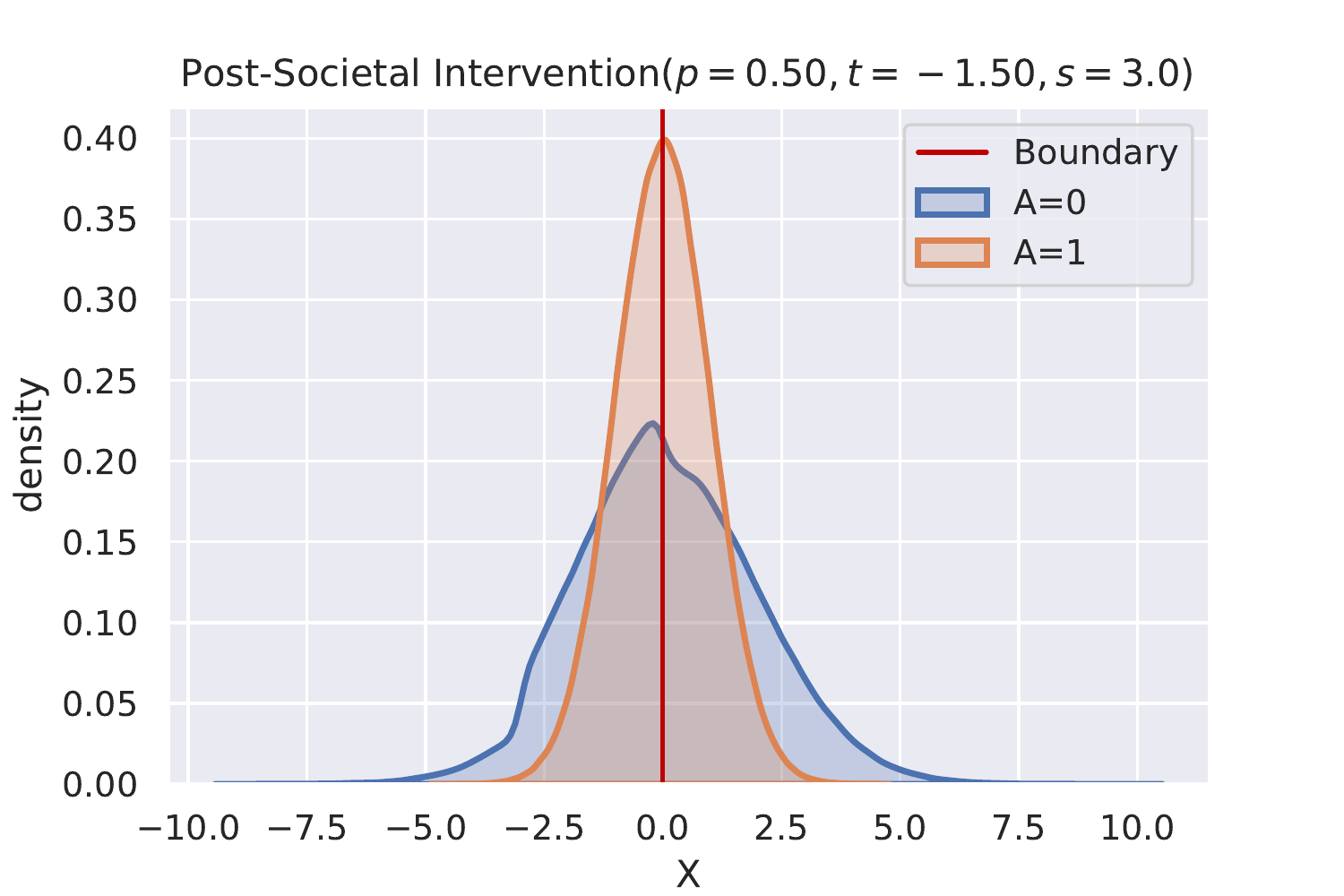}
    \end{subfigure}%
    \begin{subfigure}{0.33\textwidth}
        \includegraphics[width=\textwidth]{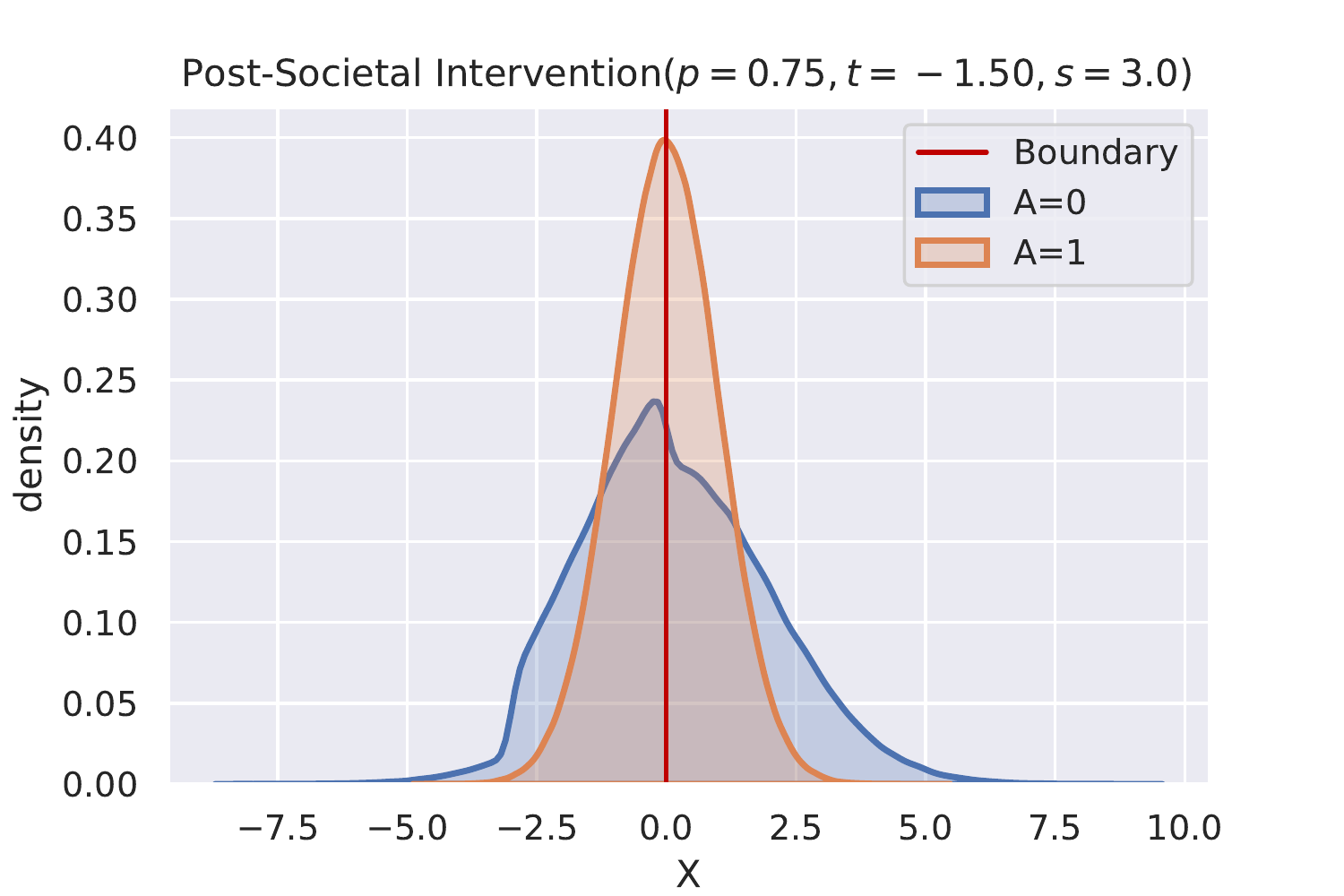}
    \end{subfigure}%
    
    \begin{subfigure}{0.33\textwidth}
        \includegraphics[width=\textwidth]{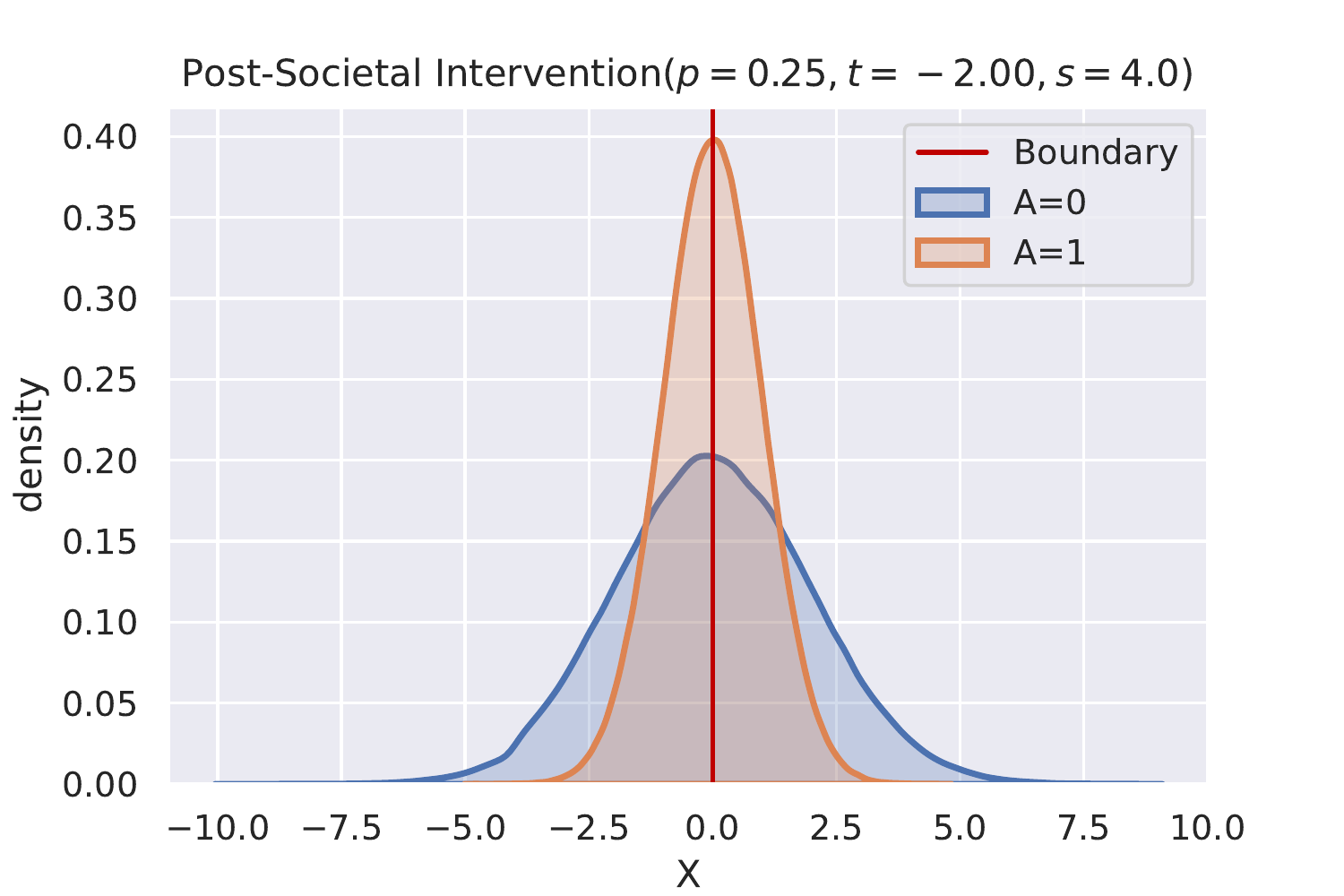}
    \end{subfigure}%
    \begin{subfigure}{0.33\textwidth}
        \includegraphics[width=\textwidth]{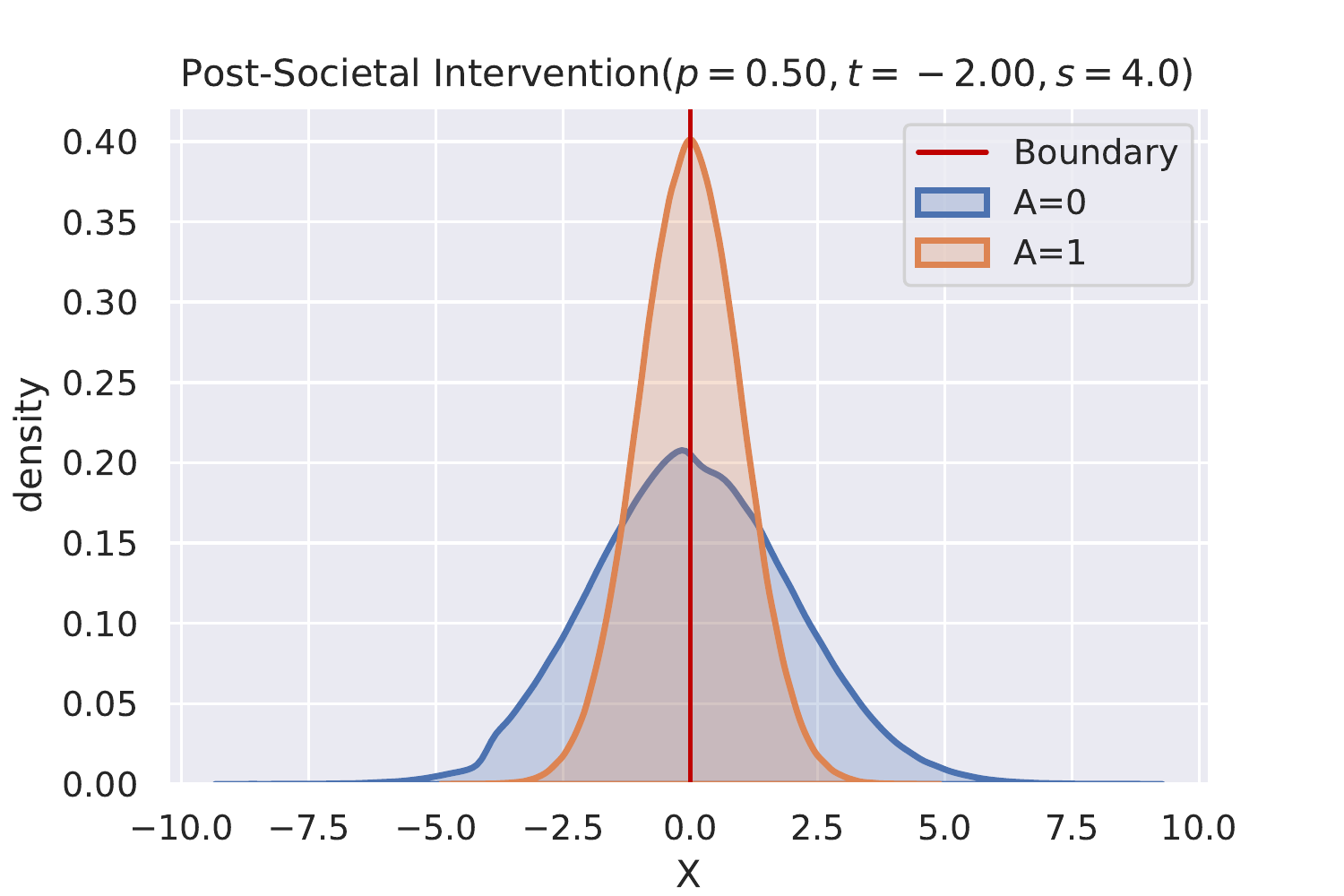}
    \end{subfigure}%
    \begin{subfigure}{0.33\textwidth}
        \includegraphics[width=\textwidth]{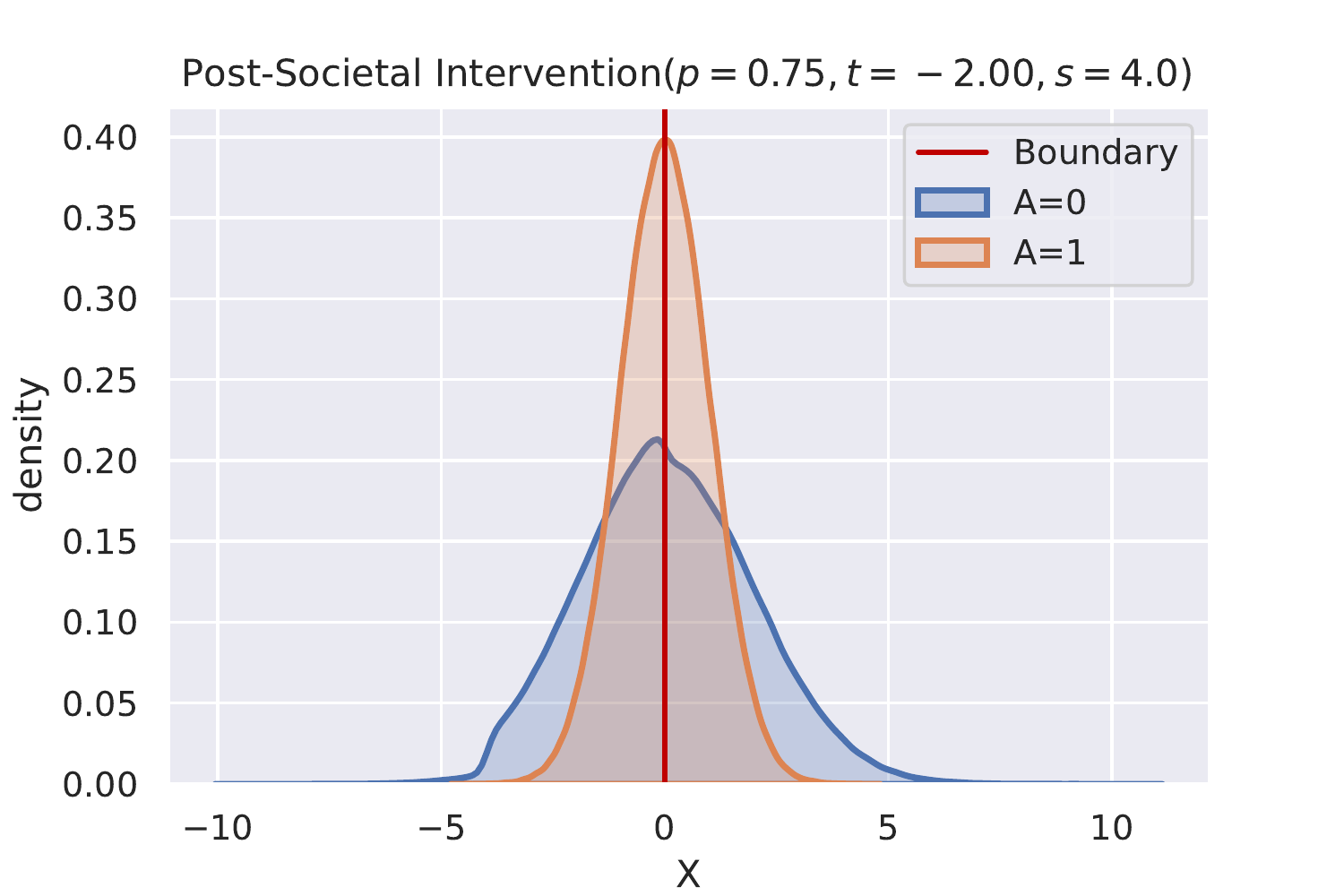}
    \end{subfigure}%
    \caption{Plots for additional societal interventions $i_k=(p, t, s)$ in the context of the credit card approval example.
    We consider budgets of $p=0.25$ (left column), $p=0.5$ (middle column), and $p=0.75$ (right column).
    In the top row, we show the difference across groups in the average distance to the decision boundary for negatively classified individuals (recourse difference), the proportion of negatively-classified individuals in the disadvantaged group $A=0$ who received the treatment (proportion treated), and the amount of subsidies actually paid out to these individuals (subsidies spent) as a function of the threshold $t$. Note that the subsidy amount is fixed to its maximum amount without affecting the label distribution, i.e., $s=-2t$.
    In rows 2-5, we show the feature distribution resulting from $i_k=(p,t,s)$ with $t=-2s\in\{-0.5, -1, -1.5, -2\}$, see the plot titles for the exact values.
    }
    \label{fig:more_societal_interventions}
\end{figure*}

In~\cref{sec:social_interventions}, we only showed plots for $i_k$ with $p=1$ since this has the largest potential to reduce recourse unfairness.
However, it may not be feasible to give subsidies to all eligible individuals, and so, for completeness, we also show plots similar to~\Cref{fig:post_intervention} for different choices of $(p,t,s)$  in~\Cref{fig:more_societal_interventions}.

\subsection{Using different SCMs for Recourse}
\label{sec:app_different_scm_estimates}
The results presented in~\cref{sec:experiments_numerical} of the main paper use an estimate $\hat{\Mcal}_\textsc{kr}$ of
the ground truth SCM $\Mcal^\star$ (learnt via kernel ridge regression under an additive noise assumption)
to solve the recourse optimisation problem.

In~\Cref{tab:appendix_different_SCM_estimates} we show a more complete account of results which also includes the cases where the ground truth SCM $\Mcal^\star$ or a linear ($\Hat{\Mcal}_\textsc{LIN}$)  
estimate thereof
is used 
as the basis for computing recourse actions.
When using an SCM estimate for recourse, we only consider \textit{valid} actions to compute $\Delta_\textbf{cost}$ and $\Delta_\textbf{ind}$, where the validity of an action is determined by whether it results in a changed prediction according the oracle $\Mcal^\star$.

We find that, as expected, using different SCMs does not affect \textbf{Acc} or $\Delta_\textbf{dist}$ since these metrics are, by definition, agnostic to the underlying causal generative process encoded by the SCM.
However, using an estimated SCM in place of the true one may result in different values for $\Delta_\textbf{cost}$ and $\Delta_\textbf{ind}$ since these metrics take the downstream effects of recourse actions on other features into account and thus depend on the underlying SCM, c.f.~\cref{def:MINT-fair,def:counterfactually-MINT-fair}.

We observe that using an estimated SCM may lead to underestimating or overestimating the true fair causal recourse metric (without any apparent clear trend as to when one or the other occurs).
Moreover, the mis-estimation of fair causal recourse metrics is particularly pronounced when using the linear SCM estimate $\Hat{\Mcal}_\textsc{LIN}$ in a scenario in which the true SCM is, in fact, nonlinear, i.e., on the CAU-ANM data sets. 
This behaviour is intuitive and to be expected and should caution against using overly strong assumptions or too simplistic parametric models when estimating an SCM for use in (fair) recourse.
We also remark that, in practice, underestimation of the true fairness metric is probably more problematic than overestimation.

Despite some small differences,
the overall trends reported in~\cref{sec:experiments_numerical} remain very much the same, and thus seem relatively robust to small differences in the SCM which is used to compute recourse actions.

\subsection{Kernel selection by cross validation}
\label{sec:app_different_kernel_choices}
For completeness, we perform the same set of experiments shown in~\Cref{tab:appendix_different_SCM_estimates} where we also choose the kernel function by cross validation, instead of fixing it to either a linear or a polynomial kernel as before.
The results are shown in~\Cref{tab:appendix_different_kernel_choices} and the overall trends, again, remain largely the same. 

As expected, we observe variations in accuracy compared to~\cref{tab:appendix_different_SCM_estimates} due to the different kernel choice. 
Perhaps most interestingly, the FairSVM seems to generally perform slightly better in terms of $\Delta_\textbf{dist}$ when given the ``free'' choice of kernel, especially on the first three data sets with linearly generated labels. This suggests that \textit{the use of a nonlinear kernel may be important for  FairSVM to achieve its goal.}

However, we caution that the results in~\Cref{tab:appendix_different_kernel_choices} may not be easily comparable across classifiers as distances are computed in the induced feature spaces which are either low-dimensional (in case of a linear kernel), high-dimensional (in case of a polynomial kernel), or infinite-dimensional (in case of an RBF kernel), which is also why we chose to report results based on the same kernel type in~\cref{sec:experiments}.

\onecolumn
\begin{landscape}
\thispagestyle{empty}
\begin{table}[hb]
    \ra{1.2}
    \centering
    \vspace{-3em}
    \caption{\textbf{Complete account of experimental results corresponding to the setting described in~\cref{sec:experiments} of the main paper, where we additionally consider using the true SCM $\Mcal^\star$ or a linear $(\Hat{\Mcal}_\text{LIN})$  estimate thereof  to infer the latent variables $\Ub$  and solve the recourse optimisation problem.}
    We compare different classifiers with respect to accuracy and different recourse fairness metrics on our three synthetic data sets with ground truth labels drawn from either a linear or a nonlinear logistic regression. 
   For ease of comparison, we use the same kernel for all SVM variants for a given dataset: a linear kernel for linearly generated ground truth labels and a polynomial kernel for non-linearly generated ground truth labels.
   Moreover, linear (resp.\ nonlinear) logistic regression classifiers are used for linearly (resp.\ nonlinearly) generated ground truth labels.
   All other hyper-parameters are chosen by 10-fold cross-validation.
   We use a dataset of 500 observations for all experiments and make sure that it is roughly balanced, both with respect to the protected attribute $A$ and the label $Y$.
   Accuracies (higher is better) are computed on a separate i.i.d.\ test set of equal size.
   Fairness metrics (lower is better) are computed based on randomly selecting 50 negatively-classified samples from each of the two protected groups and using these to compute the difference between group-wise averages ($\Delta_\textbf{dist}$ and $\Delta_\textbf{cost}$) and maximum individual unfairness. When using an SCM estimate for recourse, we only consider valid actions to compute $\Delta_\textbf{cost}$ and $\Delta_\textbf{ind}$, where the validity of an action is determined by whether it results in a changed prediction according the oracle $\Mcal^\star$. For each experiment and metric, the best performing method is highlighted in \textbf{bold}.
    }
    \label{tab:appendix_different_SCM_estimates}
    \vspace{1em}
    \resizebox{1.3\textwidth}{!}{
    \begin{tabular}{l l  c c c c | c c c c | c c c c || c c c c | c c c c | c c c c}
         \toprule
          \multirow{3}{*}{\textbf{SCM}}
         & \multirow{3}{*}{\textbf{Classifier}}
         & \multicolumn{12}{c}{\textbf{GT labels from \textit{linear} log. reg. $\rightarrow$ using \textit{linear} kernel / \textit{linear} log. reg.}} 
         & \multicolumn{12}{c}{\textbf{GT labels from \textit{nonlinear} log. reg. $\rightarrow$ using \textit{polynomial} kernel / \textit{nonlinear} log. reg.}}
         \\
         \cmidrule(r){3-14} \cmidrule(r){15-26}
         & & \multicolumn{4}{c}{\textbf{IMF}} 
         & \multicolumn{4}{c}{\textbf{CAU-LIN}}
         & \multicolumn{4}{c}{\textbf{CAU-ANM}}
         & \multicolumn{4}{c}{\textbf{IMF}} 
         & \multicolumn{4}{c}{\textbf{CAU-LIN}}
         & \multicolumn{4}{c}{\textbf{CAU-ANM}}
         \\
         \cmidrule(r){3-6} \cmidrule(r){7-10} \cmidrule(r) {11-14}
         \cmidrule(r){15-18} \cmidrule(r){19-22} \cmidrule(r) {23-26}
         & & \textbf{Acc}
         & $\Delta_\textbf{dist}$
         & $\Delta_\textbf{cost}$
         & $\Delta_\textbf{ind}$
         & \textbf{Acc}
         & $\Delta_\textbf{dist}$
         & $\Delta_\textbf{cost}$
         & $\Delta_\textbf{ind}$
         & \textbf{Acc}
         & $\Delta_\textbf{dist}$
         & $\Delta_\textbf{cost}$
         & $\Delta_\textbf{ind}$
         & \textbf{Acc}
         & $\Delta_\textbf{dist}$
         & $\Delta_\textbf{cost}$
         & $\Delta_\textbf{ind}$
         & \textbf{Acc}
         & $\Delta_\textbf{dist}$
         & $\Delta_\textbf{cost}$
         & $\Delta_\textbf{ind}$
         & \textbf{Acc}
         & $\Delta_\textbf{dist}$
         & $\Delta_\textbf{cost}$
         & $\Delta_\textbf{ind}$
         \\
\midrule
\multirow{9}{*}{$\Mcal^\star$}
    & SVM$(\Xb, A)$
        & 86.5 & 0.96 & 0.40 & 1.63
        & 89.5 & 1.18 & 0.43 & 2.11
        & \textbf{88.2} & 0.65 & \textbf{0.12} & 2.41
        & 90.8 & 0.05 & \textbf{0.00} & 1.09
        & \textbf{91.1} & 0.07 & 0.04 & 1.06
        & 90.6 & 0.04 & 0.07 & 1.40
        \\
    & LR$(\Xb, A)$
        & \textbf{86.7} & 0.48 & 0.50 & 1.91
        & 89.5 & 0.63 & 0.49 & 2.11
        & 87.7 & 0.40 & 0.22 & 2.41
        & 90.5 & 0.08 & 0.03 & 1.06
        & 90.6 & 0.09 & \textbf{0.02} & 1.00
        & 90.6 & 0.19 & 0.18 & 1.28
        \\
    & SVM$(\Xb)$
        & 86.4 & 0.99 & 0.42 & 1.80
        & 89.4 & 1.61 & 0.61 & 2.11
        & 88.0 & 0.56 & \textbf{0.12} & 2.79
        & \textbf{91.4} & 0.13 & \textbf{0.00} & 0.92
        & 91.0 & 0.17 & 0.09 & 1.09
        & \textbf{91.0} & \textbf{0.02} & \textbf{0.02} & 1.64
        \\
    & LR$(\Xb)$
        & 86.6 & 0.47 & 0.53 & 1.80
        & 89.5 & 0.64 & 0.52 & 2.11
        & 87.7 & 0.41 & 0.31 & 2.79
        & 91.0 & 0.12 & 0.03 & 1.01
        & 90.6 & 0.13 & 0.10 & 1.65
        & 90.9 & 0.08 & \textbf{0.02} & 1.16
        \\ \cmidrule(r) {2-26}
    & FairSVM$(\Xb,A)$
        & 68.1 & \textbf{0.04} & 0.28 & 1.36
        & 66.8 & 0.26 & \textbf{0.12} & 0.78
        & 66.3 & 0.25 & 0.21 & 1.50
        & 90.1 & \textbf{0.02} & \textbf{0.00} & 1.15
        & 90.7 & 0.06 & 0.04 & 1.16
        & 90.3 & 0.37 & 0.03 & 1.64
        \\ \cmidrule(r) {2-26}
    & SVM$(\Xb_{\nd(A)})$
        & 65.5 & 0.05 & 0.06 & \textbf{0.00}
        & 67.4 & \textbf{0.15} & 0.17 & \textbf{0.00}
        & 65.9 & 0.31 & 0.37 & \textbf{0.00}
        & 66.7 & 0.10 & 0.06 & \textbf{0.00}
        & 58.4 & 0.05 & 0.06 & \textbf{0.00}
        & 62.0 & 0.13 & 0.11 & \textbf{0.00}
        \\
    & LR$(\Xb_{\nd(A)})$
        & 65.3 & 0.05 & \textbf{0.05} & \textbf{0.00}
        & 67.3 & 0.18 & 0.18 & \textbf{0.00}
        & 65.6 & 0.31 & 0.31 & \textbf{0.00}
        & 64.7 & \textbf{0.02} & 0.04 & \textbf{0.00}
        & 58.4 & \textbf{0.02} & \textbf{0.02} & \textbf{0.00}
        & 61.1 & \textbf{0.02} & 0.03 & \textbf{0.00}
        \\
    & SVM$(\Xb_{\nd(A)}, \Ub_{\d(A)})$
        & 86.5 & 0.96 & 0.58 & \textbf{0.00}
        & \textbf{89.6} & 1.07 & 0.70 & \textbf{0.00}
        & 88.0 & \textbf{0.21} & 0.14 & \textbf{0.00}
        & 90.7 & \textbf{0.02} & 0.03 & \textbf{0.00}
        & \textbf{91.1} & 0.15 & 0.11 & \textbf{0.00}
        & 90.1 & 0.15 & 0.12 & \textbf{0.00}
        \\
    & LR$(\Xb_{\nd(A)}, \Ub_{\d(A)})$
        & \textbf{86.7} & 0.43 & 0.90 & \textbf{0.00}
        & 89.5 & 0.35 & 0.77 & \textbf{0.00}
        & 87.8 & 0.14 & 0.34 & \textbf{0.00}
        & 90.9 & 0.28 & 0.05 & \textbf{0.00}
        & 90.9 & 0.49 & 0.07 & \textbf{0.00}
        & 90.2 & 0.43 & 0.21 & \textbf{0.00}
        \\
\midrule
\multirow{9}{*}{$\Hat{\Mcal}_{\text{LIN}}$}
    & SVM$(\Xb, A)$
        & 86.5 & 0.96 & 0.40 & 1.63
        & 89.5 & 1.18 & 0.44 & 2.11
        & \textbf{88.2} & 0.65 & 0.30 & 3.77
        & 90.8 & 0.05 & \textbf{0.00} & 1.09
        & \textbf{91.1} & 0.07 & 0.04 & 1.06
        & 90.6 & 0.04 & 0.04 & 1.49
        \\
    & LR$(\Xb, A)$
        & \textbf{86.7} & 0.48 & 0.50 & 1.91
        & 89.5 & 0.63 & 0.51 & 2.11
        & 87.7 & 0.40 & 0.43 & 3.77
        & 90.5 & 0.08 & 0.03 & 1.06
        & 90.6 & 0.09 & \textbf{0.01} & 1.00
        & 90.6 & 0.19 & 0.20 & 1.28
        \\
    & SVM$(\Xb)$
        & 86.4 & 0.99 & 0.42 & 1.80
        & 89.4 & 1.61 & 0.61 & 2.11
        & 88.0 & 0.56 & 0.20 & 3.48
        & \textbf{91.4} & 0.13 & \textbf{0.00} & 0.92
        & 91.0 & 0.17 & 0.10 & 1.09
        & \textbf{91.0} & \textbf{0.02} & 0.03 & 1.49
        \\
    & LR$(\Xb)$
        & 86.6 & 0.47 & 0.53 & 1.80
        & 89.5 & 0.64 & 0.58 & 2.11
        & 87.7 & 0.41 & 0.55 & 3.48
        & 91.0 & 0.12 & 0.03 & 1.01
        & 90.6 & 0.13 & 0.10 & 1.65
        & 90.9 & 0.08 & 0.04 & 1.66
        \\ \cmidrule(r) {2-26}
    & FairSVM$(\Xb,A)$
        & 68.1 & \textbf{0.04} & 0.28 & 1.36
        & 66.8 & 0.26 & \textbf{0.12} & 0.78
        & 66.3 & 0.25 & 0.21 & 1.50
        & 90.1 & \textbf{0.02} & \textbf{0.00} & 1.15
        & 90.7 & 0.06 & 0.05 & 1.16
        & 90.3 & 0.37 & \textbf{0.01} & 1.64
        \\ \cmidrule(r) {2-26}
    & SVM$(\Xb_{\nd(A)})$
        & 65.5 & 0.05 & 0.06 & \textbf{0.00}
        & 67.4 & \textbf{0.15} & 0.17 & \textbf{0.00}
        & 65.9 & 0.31 & 0.37 & \textbf{0.00}
        & 66.7 & 0.10 & 0.06 & \textbf{0.00}
        & 58.4 & 0.05 & 0.06 & \textbf{0.00}
        & 62.0 & 0.13 & 0.11 & \textbf{0.00}
        \\
    & LR$(\Xb_{\nd(A)})$
        & 65.3 & 0.05 & \textbf{0.05} & \textbf{0.00}
        & 67.3 & 0.18 & 0.18 & \textbf{0.00}
        & 65.6 & 0.31 & 0.31 & \textbf{0.00}
        & 64.7 & \textbf{0.02} & 0.04 & \textbf{0.00}
        & 58.4 & \textbf{0.02} & 0.02 & \textbf{0.00}
        & 61.1 & \textbf{0.02} & 0.03 & \textbf{0.00}
        \\
    & SVM$(\Xb_{\nd(A)}, \Ub_{\d(A)})$
        & 86.5 & 0.96 & 0.58 & \textbf{0.00}
        & \textbf{89.6} & 1.07 & 0.70 & \textbf{0.00}
        & 88.0 & \textbf{0.21} & \textbf{0.14} & \textbf{0.00}
        & 90.7 & \textbf{0.02} & 0.03 & \textbf{0.00}
        & \textbf{91.1} & 0.15 & 0.11 & \textbf{0.00}
        & 90.1 & 0.15 & 0.12 & \textbf{0.00}
        \\
    & LR$(\Xb_{\nd(A)}, \Ub_{\d(A)})$
        & \textbf{86.7} & 0.43 & 0.90 & \textbf{0.00}
        & 89.5 & 0.35 & 0.77 & \textbf{0.00}
        & 87.8 & 0.14 & 0.34 & \textbf{0.00}
        & 90.9 & 0.28 & 0.05 & \textbf{0.00}
        & 90.9 & 0.49 & 0.07 & \textbf{0.00}
        & 90.2 & 0.43 & 0.21 & \textbf{0.00}
        \\
\midrule
\multirow{9}{*}{$\Hat{\Mcal}_{\text{KR}}$}
    & SVM$(\Xb, A)$
        & 86.5 & 0.96 & 0.40 & 1.63
        & 89.5 & 1.18 & 0.44 & 2.11
        & \textbf{88.2} & 0.65 & 0.27 & 2.32
        & 90.8 & 0.05 & \textbf{0.00} & 1.09
        & \textbf{91.1} & 0.07 & 0.03 & 1.06
        & 90.6 & 0.04 & 0.03 & 1.40
        \\
    & LR$(\Xb, A)$
        & \textbf{86.7} & 0.48 & 0.50 & 1.91
        & 89.5 & 0.63 & 0.53 & 2.11
        & 87.7 & 0.40 & 0.34 & 2.32
        & 90.5 & 0.08 & 0.03 & 1.06
        & 90.6 & 0.09 & \textbf{0.01} & 1.00
        & 90.6 & 0.19 & 0.22 & 1.28
        \\
    & SVM$(\Xb)$
        & 86.4 & 0.99 & 0.42 & 1.80
        & 89.4 & 1.61 & 0.61 & 2.11
        & 88.0 & 0.56 & 0.29 & 2.79
        & \textbf{91.4} & 0.13 & \textbf{0.00} & 0.92
        & 91.0 & 0.17 & 0.08 & 1.09
        & \textbf{91.0} & \textbf{0.02} & 0.03 & 1.64
        \\
    & LR$(\Xb)$
        & 86.6 & 0.47 & 0.53 & 1.80
        & 89.5 & 0.64 & 0.57 & 2.11
        & 87.7 & 0.41 & 0.43 & 2.79
        & 91.0 & 0.12 & 0.03 & 1.01
        & 90.6 & 0.13 & 0.10 & 1.65
        & 90.9 & 0.08 & 0.06 & 1.66
        \\ \cmidrule(r) {2-26}
    & FairSVM$(\Xb,A)$
        & 68.1 & \textbf{0.04} & 0.28 & 1.36
        & 66.8 & 0.26 & \textbf{0.12} & 0.78
        & 66.3 & 0.25 & 0.21 & 1.50
        & 90.1 & \textbf{0.02} & \textbf{0.00} & 1.15
        & 90.7 & 0.06 & 0.04 & 1.16
        & 90.3 & 0.37 & \textbf{0.02} & 1.64
        \\ \cmidrule(r) {2-26}
    & SVM$(\Xb_{\nd(A)})$
        & 65.5 & 0.05 & 0.06 & \textbf{0.00}
        & 67.4 & \textbf{0.15} & 0.17 & \textbf{0.00}
        & 65.9 & 0.31 & 0.37 & \textbf{0.00}
        & 66.7 & 0.10 & 0.06 & \textbf{0.00}
        & 58.4 & 0.05 & 0.06 & \textbf{0.00}
        & 62.0 & 0.13 & 0.11 & \textbf{0.00}
        \\
    & LR$(\Xb_{\nd(A)})$
        & 65.3 & 0.05 & \textbf{0.05} & \textbf{0.00}
        & 67.3 & 0.18 & 0.18 & \textbf{0.00}
        & 65.6 & 0.31 & 0.31 & \textbf{0.00}
        & 64.7 & \textbf{0.02} & 0.04 & \textbf{0.00}
        & 58.4 & \textbf{0.02} & 0.02 & \textbf{0.00}
        & 61.1 & \textbf{0.02} & 0.03 & \textbf{0.00}
        \\
    & SVM$(\Xb_{\nd(A)}, \Ub_{\d(A)})$
        & 86.5 & 0.96 & 0.58 & \textbf{0.00}
        & \textbf{89.6} & 1.07 & 0.70 & \textbf{0.00}
        & 88.0 & \textbf{0.21} & \textbf{0.14} & \textbf{0.00}
        & 90.7 & \textbf{0.02} & 0.03 & \textbf{0.00}
        & \textbf{91.1} & 0.15 & 0.11 & \textbf{0.00}
        & 90.1 & 0.15 & 0.12 & \textbf{0.00}
        \\
    & LR$(\Xb_{\nd(A)}, \Ub_{\d(A)})$
        & \textbf{86.7} & 0.43 & 0.90 & \textbf{0.00}
        & 89.5 & 0.35 & 0.77 & \textbf{0.00}
        & 87.8 & 0.14 & 0.34 & \textbf{0.00}
        & 90.9 & 0.28 & 0.05 & \textbf{0.00}
        & 90.9 & 0.49 & 0.07 & \textbf{0.00}
        & 90.2 & 0.43 & 0.21 & \textbf{0.00}
        \\
\bottomrule 
    \end{tabular}
    }
\end{table}
\centering

\clearpage
\thispagestyle{empty}
\begin{table}[hb]
    \ra{1.2}
    \centering
    \caption{\textbf{Additional results where also the kernel (linear, polynomial, or rbf) for each SVM is chosen by 5-fold cross-validation instead of being fixed based on the ground truth label distribution.}
    We remark that some metrics (e.g., $\Delta_\textbf{dist}$) may not be comparable across methods since they are computed in a different reference space when different kernels are selected.
    Otherwise the experimental setup is identical to that from~\cref{tab:appendix_different_SCM_estimates}, see the caption for details.
    }
    \label{tab:appendix_different_kernel_choices}
    \vspace{1em}
    \resizebox{1.3\textwidth}{!}{
    \begin{tabular}{l l  c c c c | c c c c | c c c c || c c c c | c c c c | c c c c}
         \toprule
          \multirow{3}{*}{\textbf{SCM}}
         & \multirow{3}{*}{\textbf{Classifier}}
         & \multicolumn{12}{c}{\textbf{GT labels from \textit{linear} log. reg. $\rightarrow$ using \textit{cross-validated} kernel / \textit{linear} log. reg.}} 
        & \multicolumn{12}{c}{\textbf{GT labels from \textit{nonlinear} log. reg. $\rightarrow$ using \textit{cross-validated} kernel / \textit{nonlinear} log. reg.}}
         \\
         \cmidrule(r){3-14} \cmidrule(r){15-26}
         & & \multicolumn{4}{c}{\textbf{IMF}} 
         & \multicolumn{4}{c}{\textbf{CAU-LIN}}
         & \multicolumn{4}{c}{\textbf{CAU-ANM}}
         & \multicolumn{4}{c}{\textbf{IMF}} 
         & \multicolumn{4}{c}{\textbf{CAU-LIN}}
         & \multicolumn{4}{c}{\textbf{CAU-ANM}}
         \\
         \cmidrule(r){3-6} \cmidrule(r){7-10} \cmidrule(r) {11-14}
         \cmidrule(r){15-18} \cmidrule(r){19-22} \cmidrule(r) {23-26}
         & & \textbf{Acc}
         & $\Delta_\textbf{dist}$
         & $\Delta_\textbf{cost}$
         & $\Delta_\textbf{ind}$
         & \textbf{Acc}
         & $\Delta_\textbf{dist}$
         & $\Delta_\textbf{cost}$
         & $\Delta_\textbf{ind}$
         & \textbf{Acc}
         & $\Delta_\textbf{dist}$
         & $\Delta_\textbf{cost}$
         & $\Delta_\textbf{ind}$
         & \textbf{Acc}
         & $\Delta_\textbf{dist}$
         & $\Delta_\textbf{cost}$
         & $\Delta_\textbf{ind}$
         & \textbf{Acc}
         & $\Delta_\textbf{dist}$
         & $\Delta_\textbf{cost}$
         & $\Delta_\textbf{ind}$
         & \textbf{Acc}
         & $\Delta_\textbf{dist}$
         & $\Delta_\textbf{cost}$
         & $\Delta_\textbf{ind}$
         \\
\midrule
\multirow{9}{*}{$\Mcal^\star$}
    & SVM$(\Xb, A)$
        & 86.5 & 0.96 & 0.40 & 1.63
        & 89.2 & 1.33 & 0.55 & 2.10
        & \textbf{87.8} & 0.36 & \textbf{0.08} & 2.79
        & 90.8 & 0.05 & \textbf{0.00} & 1.09
        & \textbf{91.1} & 0.07 & 0.04 & 1.06
        & 90.6 & 0.04 & 0.07 & 1.40
        \\
    & LR$(\Xb, A)$
        & \textbf{86.7} & 0.48 & 0.50 & 1.91
        & \textbf{89.5} & 0.63 & 0.49 & 2.11
        & 87.7 & 0.40 & 0.22 & 2.41
        & 90.5 & 0.08 & 0.03 & 1.06
        & 90.6 & 0.09 & \textbf{0.02} & 1.00
        & 90.6 & 0.19 & 0.18 & 1.28
        \\
    & SVM$(\Xb)$
        & 86.4 & 0.99 & 0.42 & 1.80
        & \textbf{89.5} & 1.13 & 0.53 & 2.14
        & 87.6 & 0.43 & 0.42 & 2.79
        & \textbf{91.4} & 0.13 & \textbf{0.00} & 0.92
        & 91.0 & 0.17 & 0.09 & 1.09
        & 89.4 & 0.16 & 0.16 & 1.16
        \\
    & LR$(\Xb)$
        & 86.6 & 0.47 & 0.53 & 1.80
        & \textbf{89.5} & 0.64 & 0.52 & 2.11
        & 87.7 & 0.41 & 0.31 & 2.79
        & 91.0 & 0.12 & 0.03 & 1.01
        & 90.6 & 0.13 & 0.10 & 1.65
        & \textbf{90.9} & 0.08 & \textbf{0.02} & 1.16
        \\ \cmidrule(r) {2-26}
    & FairSVM$(\Xb,A)$
        & 86.4 & \textbf{0.01} & 0.20 & 1.61
        & 60.5 & \textbf{0.00} & 0.33 & 1.05
        & 57.6 & \textbf{0.01} & 0.13 & 1.76
        & 90.1 & 0.02 & \textbf{0.00} & 1.15
        & 90.7 & 0.06 & 0.04 & 1.16
        & 78.0 & \textbf{0.00} & 0.04 & 1.73
        \\ \cmidrule(r) {2-26}
    & SVM$(\Xb_{\nd(A)})$
        & 64.6 & 0.06 & 0.09 & \textbf{0.00}
        & 67.3 & 0.17 & 0.25 & \textbf{0.00}
        & 65.9 & 0.28 & 0.31 & \textbf{0.00}
        & 65.7 & \textbf{0.01} & 0.02 & \textbf{0.00}
        & 55.6 & 0.04 & 0.03 & \textbf{0.00}
        & 61.6 & 0.04 & 0.03 & \textbf{0.00}
        \\
    & LR$(\Xb_{\nd(A)})$
        & 65.3 & 0.05 & \textbf{0.05} & \textbf{0.00}
        & 67.3 & 0.18 & \textbf{0.18} & \textbf{0.00}
        & 65.6 & 0.31 & 0.31 & \textbf{0.00}
        & 64.7 & 0.02 & 0.04 & \textbf{0.00}
        & 58.4 & \textbf{0.02} & \textbf{0.02} & \textbf{0.00}
        & 61.1 & 0.02 & 0.03 & \textbf{0.00}
        \\
    & SVM$(\Xb_{\nd(A)}, \Ub_{\d(A)})$
        & 86.6 & 0.84 & 0.64 & \textbf{0.00}
        & 89.4 & 0.81 & 0.54 & \textbf{0.00}
        & 87.4 & 0.21 & 0.35 & \textbf{0.00}
        & 90.7 & 0.02 & 0.03 & \textbf{0.00}
        & \textbf{91.1} & 0.15 & 0.11 & \textbf{0.00}
        & 89.0 & 0.31 & 0.13 & \textbf{0.00}
        \\
    & LR$(\Xb_{\nd(A)}, \Ub_{\d(A)})$
        & \textbf{86.7} & 0.43 & 0.90 & \textbf{0.00}
        & \textbf{89.5} & 0.35 & 0.77 & \textbf{0.00}
        & \textbf{87.8} & 0.14 & 0.34 & \textbf{0.00}
        & 90.9 & 0.28 & 0.05 & \textbf{0.00}
        & 90.9 & 0.49 & 0.07 & \textbf{0.00}
        & 90.2 & 0.43 & 0.21 & \textbf{0.00}
        \\
\midrule
\multirow{9}{*}{$\Hat{\Mcal}_{\text{LIN}}$}
    & SVM$(\Xb, A)$
        & 86.5 & 0.96 & 0.40 & 1.63
        & 89.2 & 1.33 & 0.55 & 2.10
        & \textbf{87.8} & 0.36 & 0.13 & 3.48
        & 90.8 & 0.05 & \textbf{0.00} & 1.09
        & \textbf{91.1} & 0.07 & 0.04 & 1.06
        & 90.6 & 0.04 & 0.04 & 1.49
        \\
    & LR$(\Xb, A)$
        & \textbf{86.7} & 0.48 & 0.50 & 1.91
        & \textbf{89.5} & 0.63 & 0.51 & 2.11
        & 87.7 & 0.40 & 0.43 & 3.77
        & 90.5 & 0.08 & 0.03 & 1.06
        & 90.6 & 0.09 & \textbf{0.01} & 1.00
        & 90.6 & 0.19 & 0.20 & 1.28
        \\
    & SVM$(\Xb)$
        & 86.4 & 0.99 & 0.42 & 1.80
        & \textbf{89.5} & 1.13 & 0.51 & 2.14
        & 87.6 & 0.43 & 0.42 & 4.05
        & \textbf{91.4} & 0.13 & \textbf{0.00} & 0.92
        & 91.0 & 0.17 & 0.10 & 1.09
        & 89.4 & 0.16 & 0.11 & 1.16
        \\
    & LR$(\Xb)$
        & 86.6 & 0.47 & 0.53 & 1.80
        & \textbf{89.5} & 0.64 & 0.58 & 2.11
        & 87.7 & 0.41 & 0.55 & 3.48
        & 91.0 & 0.12 & 0.03 & 1.01
        & 90.6 & 0.13 & 0.10 & 1.65
        & \textbf{90.9} & 0.08 & 0.04 & 1.66
        \\ \cmidrule(r) {2-26}
    & FairSVM$(\Xb,A)$
        & 86.4 & \textbf{0.01} & 0.20 & 1.61
        & 60.5 & \textbf{0.00} & 0.29 & 1.05
        & 57.6 & \textbf{0.01} & \textbf{0.12} & 1.76
        & 90.1 & 0.02 & \textbf{0.00} & 1.15
        & 90.7 & 0.06 & 0.05 & 1.16
        & 78.0 & \textbf{0.00} & \textbf{0.03} & 1.73
        \\ \cmidrule(r) {2-26}
    & SVM$(\Xb_{\nd(A)})$
        & 64.6 & 0.06 & 0.09 & \textbf{0.00}
        & 67.3 & 0.17 & 0.25 & \textbf{0.00}
        & 65.9 & 0.28 & 0.31 & \textbf{0.00}
        & 65.7 & \textbf{0.01} & 0.02 & \textbf{0.00}
        & 55.6 & 0.04 & 0.03 & \textbf{0.00}
        & 61.6 & 0.04 & \textbf{0.03} & \textbf{0.00}
        \\
    & LR$(\Xb_{\nd(A)})$
        & 65.3 & 0.05 & \textbf{0.05} & \textbf{0.00}
        & 67.3 & 0.18 & \textbf{0.18} & \textbf{0.00}
        & 65.6 & 0.31 & 0.31 & \textbf{0.00}
        & 64.7 & 0.02 & 0.04 & \textbf{0.00}
        & 58.4 & \textbf{0.02} & 0.02 & \textbf{0.00}
        & 61.1 & 0.02 & \textbf{0.03} & \textbf{0.00}
        \\
    & SVM$(\Xb_{\nd(A)}, \Ub_{\d(A)})$
        & 86.6 & 0.84 & 0.64 & \textbf{0.00}
        & 89.4 & 0.81 & 0.54 & \textbf{0.00}
        & 87.4 & 0.21 & 0.35 & \textbf{0.00}
        & 90.7 & 0.02 & 0.03 & \textbf{0.00}
        & \textbf{91.1} & 0.15 & 0.11 & \textbf{0.00}
        & 89.0 & 0.31 & 0.13 & \textbf{0.00}
        \\
    & LR$(\Xb_{\nd(A)}, \Ub_{\d(A)})$
        & \textbf{86.7} & 0.43 & 0.90 & \textbf{0.00}
        & \textbf{89.5} & 0.35 & 0.77 & \textbf{0.00}
        & \textbf{87.8} & 0.14 & 0.34 & \textbf{0.00}
        & 90.9 & 0.28 & 0.05 & \textbf{0.00}
        & 90.9 & 0.49 & 0.07 & \textbf{0.00}
        & 90.2 & 0.43 & 0.21 & \textbf{0.00}
        \\
\midrule
\multirow{9}{*}{$\Hat{\Mcal}_{\text{KR}}$}
    & SVM$(\Xb, A)$
        & 86.5 & 0.96 & 0.40 & 1.63
        & 89.2 & 1.33 & 0.56 & 2.10
        & \textbf{87.8} & 0.36 & 0.18 & 2.79
        & 90.8 & 0.05 & \textbf{0.00} & 1.09
        & \textbf{91.1} & 0.07 & 0.03 & 1.06
        & 90.6 & 0.04 & 0.03 & 1.40
        \\
    & LR$(\Xb, A)$
        & \textbf{86.7} & 0.48 & 0.50 & 1.91
        & \textbf{89.5} & 0.63 & 0.53 & 2.11
        & 87.7 & 0.40 & 0.34 & 2.32
        & 90.5 & 0.08 & 0.03 & 1.06
        & 90.6 & 0.09 & \textbf{0.01} & 1.00
        & 90.6 & 0.19 & 0.22 & 1.28
        \\
    & SVM$(\Xb)$
        & 86.4 & 0.99 & 0.42 & 1.80
        & \textbf{89.5} & 1.13 & 0.52 & 2.14
        & 87.6 & 0.43 & 0.44 & 2.79
        & \textbf{91.4} & 0.13 & \textbf{0.00} & 0.92
        & 91.0 & 0.17 & 0.08 & 1.09
        & 89.4 & 0.16 & 0.14 & 1.16
        \\
    & LR$(\Xb)$
        & 86.6 & 0.47 & 0.53 & 1.80
        & \textbf{89.5} & 0.64 & 0.57 & 2.11
        & 87.7 & 0.41 & 0.43 & 2.79
        & 91.0 & 0.12 & 0.03 & 1.01
        & 90.6 & 0.13 & 0.10 & 1.65
        & \textbf{90.9} & 0.08 & 0.06 & 1.66
        \\ \cmidrule(r) {2-26}
    & FairSVM$(\Xb,A)$
        & 86.4 & \textbf{0.01} & 0.20 & 1.61
        & 60.5 & \textbf{0.00} & 0.26 & 1.50
        & 57.6 & \textbf{0.01} & \textbf{0.12} & 1.76
        & 90.1 & 0.02 & \textbf{0.00} & 1.15
        & 90.7 & 0.06 & 0.04 & 1.16
        & 78.0 & \textbf{0.00} & \textbf{0.01} & 1.73
        \\ \cmidrule(r) {2-26}
    & SVM$(\Xb_{\nd(A)})$
        & 64.6 & 0.06 & 0.09 & \textbf{0.00}
        & 67.3 & 0.17 & 0.25 & \textbf{0.00}
        & 65.9 & 0.28 & 0.31 & \textbf{0.00}
        & 65.7 & \textbf{0.01} & 0.02 & \textbf{0.00}
        & 55.6 & 0.04 & 0.03 & \textbf{0.00}
        & 61.6 & 0.04 & 0.03 & \textbf{0.00}
        \\
    & LR$(\Xb_{\nd(A)})$
        & 65.3 & 0.05 & \textbf{0.05} & \textbf{0.00}
        & 67.3 & 0.18 & \textbf{0.18} & \textbf{0.00}
        & 65.6 & 0.31 & 0.31 & \textbf{0.00}
        & 64.7 & 0.02 & 0.04 & \textbf{0.00}
        & 58.4 & \textbf{0.02} & 0.02 & \textbf{0.00}
        & 61.1 & 0.02 & 0.03 & \textbf{0.00}
        \\
    & SVM$(\Xb_{\nd(A)}, \Ub_{\d(A)})$
        & 86.6 & 0.84 & 0.64 & \textbf{0.00}
        & 89.4 & 0.81 & 0.54 & \textbf{0.00}
        & 87.4 & 0.21 & 0.35 & \textbf{0.00}
        & 90.7 & 0.02 & 0.03 & \textbf{0.00}
        & \textbf{91.1} & 0.15 & 0.11 & \textbf{0.00}
        & 89.0 & 0.31 & 0.13 & \textbf{0.00}
        \\
    & LR$(\Xb_{\nd(A)}, \Ub_{\d(A)})$
        & \textbf{86.7} & 0.43 & 0.90 & \textbf{0.00}
        & \textbf{89.5} & 0.35 & 0.77 & \textbf{0.00}
        & \textbf{87.8} & 0.14 & 0.34 & \textbf{0.00}
        & 90.9 & 0.28 & 0.05 & \textbf{0.00}
        & 90.9 & 0.49 & 0.07 & \textbf{0.00}
        & 90.2 & 0.43 & 0.21 & \textbf{0.00}
        \\
\bottomrule 
    \end{tabular}
    }
\end{table}

\end{landscape}

\end{document}